\documentclass[lettersize,journal]{IEEEtran}
\usepackage{amsmath,amsfonts}
\usepackage{algorithmic}
\usepackage{array}
\usepackage[caption=false,font=small,labelfont=rm,textfont=rm]{subfig}
\usepackage{textcomp}
\usepackage{stfloats}
\usepackage{url}
\usepackage{verbatim}
\usepackage{graphicx}
\def\BibTeX{{\rm B\kern-.05em{\sc i\kern-.025em b}\kern-.08em
		T\kern-.1667em\lower.7ex\hbox{E}\kern-.125emX}}
\usepackage{balance}
\usepackage[utf8]{inputenc}
\usepackage[linesnumbered,ruled,vlined]{algorithm2e}
\usepackage{hyperref}
\usepackage[switch]{lineno}
\usepackage{cite}

\hypersetup{
	colorlinks = true, 
	linkcolor = blue,
	urlcolor = blue,
	citecolor = blue
}
\begin{document}
	
	\title{DRL-Based Resource Allocation for Motion Blur Resistant Federated Self-Supervised Learning in IoV
	}
	
	\author{
		
		{
			Xueying Gu,
			Qiong Wu,~\IEEEmembership{Senior Member,~IEEE}, 
			Pingyi Fan,~\IEEEmembership{Senior Member,~IEEE}, 
			Qiang Fan,\\
			Nan Cheng,~\IEEEmembership{Senior Member,~IEEE}, 
			Wen Chen,~\IEEEmembership{Senior Member,~IEEE}, 
			 and Khaled B. Letaief,~\IEEEmembership{Fellow,~IEEE}

		}
		
		\thanks{
			
			{This work was supported in part by the National Natural Science Foundation of China under Grant 61701197 and Grant 62071296; in part by the National Key Research and Development Program of China under Grant 2021YFA1000500(4); in part by the National Key Project under Grant 2020YFB1807700; in part by Shanghai Kewei under Grant 22JC1404000; in part by the Research Grants Council under the Areas of Excellence Scheme under Grant AoE/E-601/22-R; and in part by the 111 Project under Grant B23008. \emph{(Corresponding author: Qiong Wu.)}
				
			Xueying Gu and Qiong Wu are with the School of Internet of Things Engineering, Jiangnan University, Wuxi 214122, China (e-mail: xueyinggu@stu.jiangnan.edu.cn, qiongwu@jiangnan.edu.cn)
				
			Pingyi Fan is with the Department of Electronic Engineering, Beijing National Research Center for Information Science and Technology, Tsinghua University, Beijing 100084, China (e-mail: fpy@tsinghua.edu.cn)
			
			Qiang Fan is with Qualcomm, San Jose, CA 95110, USA (e-mail: qf9898@gmail.com)
			
			Nan Cheng is with the State Key Lab. of ISN and School of Telecommunications Engineering, Xidian University, Xi’an 710071, China (e-mail: dr.nan.cheng@ieee.org)
			
			Wen Chen is with the Department of Electronic Engineering, Shanghai Jiao Tong University, Shanghai 200240, China (e-mail: wenchen@sjtu.edu.cn)
			
			K. B. Letaief is with the Department of Electrical and Computer Engineering, the Hong Kong University of Science and Technology (HKUST), Hong Kong (email:eekhaled@ust.hk).
				
			}
			
		}
	}
	\maketitle
	
	\begin{abstract}
		In the Internet of Vehicles (IoV), Federated Learning (FL) provides a privacy-preserving solution by aggregating local models without sharing data. Traditional supervised learning requires image data with labels, but data labeling involves significant manual effort. Federated Self-Supervised Learning (FSSL) utilizes Self-Supervised Learning (SSL) for local training in FL, eliminating the need for labels while protecting privacy. Compared to other SSL methods, Momentum Contrast (MoCo) reduces the demand for computing resources and storage space by creating a dictionary. However, using MoCo in FSSL requires uploading the local dictionary from vehicles to Base Station (BS), which poses a risk of privacy leakage. Simplified Contrast (SimCo) addresses the privacy leakage issue in MoCo-based FSSL by using dual temperature instead of a dictionary to control sample distribution. Additionally, considering the negative impact of motion blur on model aggregation, and based on SimCo, we propose a motion blur-resistant FSSL method, referred to as BFSSL. Furthermore, we address energy consumption and delay in the BFSSL process by proposing a Deep Reinforcement Learning (DRL)-based resource allocation scheme, called DRL-BFSSL. In this scheme, BS allocates the Central Processing Unit (CPU) frequency and transmission power of vehicles to minimize energy consumption and latency, while aggregating received models based on the motion blur level. Simulation results validate the effectiveness of our proposed aggregation and resource allocation methods.
		
	\end{abstract}
	
	
	\section{Introduction}
	Internet of Vehicles (IoV) is a technology that connects vehicles, road and transportation facilities to Internet, enabling intelligent interaction and information sharing among them. Advances in IoV have made many practical applications possible, e.g., automatic navigation, meteorological information and road condition monitoring, which provide convenience to drivers and self-driving system, reduce the probability of accidents and improve the efficiency of the transportation system \cite{qwu2023,kzhang2015,Space/Aerial}. These applications need robust models, which require a lot of data for training. In IoV, vehicles can continuously capture the fresh image data along the moving road, including road conditions and environmental states through onboard sensors, where most of the data are captured in the form of images \cite{gluo2024,JinglongShen,KnowledgeDriven}. Moreover, vehicles also can store the image data locally and deal with them to realize recognition and classification, which provides essential information for self-driving and driver assistance systems to aid drivers in perceiving and understanding surroundings. For instance, a brand-new concept of underwater teleportation was proposed in \cite{jyang2024}, which can significantly improve the control stability and perceiving transparency of drivers via the human-on-the-loop system. 
	
	However, the locally stored image data, denoted as local data, is usually private for vehicles, and vehicles are reluctant to share the data with each other to protect privacy. Since vehicles come from all directions, the local data they collect vary greatly. If the model, denoted as local model, is trained locally based on the local data, it may converge to a local optimum \cite{CharacterizingInternet,CharacterizingUser}. In contrast to distributed training, Federated Learning (FL) performs a centered training and enables a Base Station (BS) to obtain a global model through aggregating local models from vehicles without sharing the local data, thereby providing a privacy-preserving solution and reducing the differences between local models \cite{dye2020,cmi2021}. For local training, the traditional supervised learning requires the image data with labels \cite{vgupta2023,DeepReinforcementLearningBasedAoIAwareResource,ReconfigurableIntelligentSurfaceAidedVehicularEdgeComputing}, but data labeling requires significant manual effort. Especially in IoV, a large amount of unlabeled data is generated daily, further increasing the cost of manual labeling. The Self-Supervised Learning (SSL) leverages pretext to provide training based on the inherent properties of data, thus it can learn feature representations from the data itself without the need for manually annotated labels. The traditional SSL methods obtain a better training performance through generating and storing abundant negative samples, which occupies a lot of computing resource and storage space. Thus, traditional methods are not suitable for the resource and space limited IoV. The famous SSL method Momentum Contrast (MoCo) uses $k$ values to form a dictionary and stand for the reference of negative samples, which can save computing resource and storage space \cite{khe2020}. What's more, MoCo uses dynamic dictionaries and momentum encoders to maintain a large and consistent set of negative samples, thereby improving the stability and quality of learned representations. However, employing MoCo in FSSL for local training will upload $k$ values to create a big queue in BS for aggregation, which may cause privacy leakage \cite{jzhao2021Hotfed}. Moreover, in IoV, the high-velocity movement of vehicles often leads to motion blur. Specifically, excessive vehicle velocity may result in insufficient exposure time of the camera sensor, causing image blur \cite{mzhao2019,pbinnar2016,Vehicle Selection for C-V2X Mode 4}. Motion blur significantly reduces image clarity, making the features extracted from such images less accurate and leading to unstable feature representation during model training. At the same time, critical information in blurred images may be obscured, making it difficult for feature extraction algorithms to recognize and differentiate important image content, thus impacting the model’s accuracy and reliability. In FL, local models may perform poorly due to blur in the local data, thereby affecting the performance of the global model after aggregating. Data collected from different vehicles may exhibit significant distribution differences due to variations in environmental and motion conditions. This data heterogeneity usually leads lead to performance imbalance of the model across different vehicles, further exacerbating the challenges in model training and aggregation. To the best of our knowledge, there is no work considering the privacy leakage and motion blur for FSSL in IoV. 
	
	The improved MoCo method Simplified Contrast (SimCo) uses dual temperature to control sample distribution to eliminate the usage of dictionary \cite{czhang2022dual temp}, thus it can greatly protect privacy and address the shortcomings of FedCo and MoCo in their integration with FL in IoV. 
	Considering the negative impact of motion blur on model aggregation, we also propose a motion Blur resistant FSSL, refer to as BFSSL.
	
	It is noteworthy that increasing the Central Processing Unit (CPU) frequency of a vehicle results in more local iterations, which leads to higher accuracy of classification, while consuming more energy consumption. Higher transmission power of a vehicle reduces the transmission delay, while costing more energy. Simultaneously, low CPU frequency reduces the local iterations and the classification accuracy of the global model \cite{wzhuang2019}\cite{wwang2024}. Additionally, low transmission power increases data interference, resulting in higher rates of transmission failure. Thus, it is critical to find a resource allocation scheme that allocates resources (i.e., CPU frequency and transmission power of vehicles) to minimize energy consumption and delay for the BFSSL, while increasing the success rate of transmission in IoV. Since the resource allocation problem is usually Non-convex in the complicated IoV, the traditional convex optimization methods are difficult to solve it. Deep Reinforcement Learning (DRL) is a good option to solve the Non-convex problem. In this paper, we propose the DRL-BFSSL algorithm, where vehicles employs SimCo to perform locally and sends local models to BS for blur-based aggregation. Meanwhile, BS allocates CPU frequency and transmission power to minimize the energy consumption and delay\footnote{The source code has been released at: https://github.com/qiongwu86/DRL-BFSSL}.
	
	Overall, the main contributions of this paper can be summarized as follows:
	
	\begin{itemize}
	\item Based on the concept of FL, we enable vehicles to perform SimCo locally with their own unlabeled image data. This approach not only effectively safeguards the privacy of vehicles but also avoids labeled data.
	
	\item In IoV, we utilize the level of motion blur as a crucial weighting factor on Independent and Identically Distributed (IID) and Non-IID datasets during model aggregation. Subsequently, aggregating models uploaded by training vehicles based on the blur level, we obtain a blur resistant global model. 
	
	\item  We firstly formulate an optimization problem aiming to minimize the sum of energy consumption and delay. Subsequently, utilizing the Karush-Kuhn-Tucker (KKT) conditions to determine the optimal solution that jointly allocating bandwidth and CPU frequency. Finally, by the Soft Actor-Critic (SAC) algorithm, we identify the optimal allocation schedule for the CPU frequency and transmission power of each vehicle. 
	
	\item Given that low transmission power can result in data transmission failures, we use the data error rate to establish a minimum transmission power limit.

	\end{itemize}
	
	The rest of this paper is divided into several interrelated parts to comprehensively explore the discussed problem. In Section \ref{sec2}, we will review prior work relevant to this study, providing background and context for subsequent discussions. In Section \ref{sec3}, we describe the system model. In Section \ref{sec4}, we will formulate an optimization problem aiming to minimize the sum of energy consumption and delay and in Section \ref{sec5}, we will introduce our proposed DRL-BFSSL algorithm. Section \ref{sec6} showcase and discuss the simulation results. Ultimately, we will summarize the paper and  in Section \ref{sec7}.
	
	\section{Related Work}
	\label{sec2}
	
	In this section, we will introduce the related work about FSSL and resource allocation based on DRL in IoV.
	
	\subsection{Federated SSL}
	In \cite{njahan2023}, Jahan \emph{et al.} proposed a FSSL which combines FL and the SSL method (i.e., Simple framework for Contrastive Learning of Representations (SimCLR)), to identify Mpox from skin lesion images \cite{tchen2020SimCLR}. In \cite{ryan2023}, Yan \emph{et al.} proposed a FSSL which combines FL and the SSL method (i.e., Variational inference with adversarial learning for end-to-end text-to-speech (Vits)), and utilizes medical images stored in each hospital to train models. In \cite{yxiao2013}, Xiao \emph{et al.} proposed a FSSL which integrates Generative Adversarial Networks (GAN) as a pretext for SSL and combines it with FL to support automatic traffic analysis and synthesis over a large number of heterogeneous datasets. In \cite{sli2023}, Li \emph{et al.} proposed a FSSL which integrates You Only Look Once (YOLO) as a pretext for SSL and combines it with FL to process object detection in power operation sites. These FSSL methods mentioned above require a large number of samples for training, which imposes high demands on the computing and storage capabilities of devices. However, the computing and storage resources of vehicles are limited in IoV, thus these methods are not suitable for IoV \cite{JMX}.
	
	To mitigate the demand on computing and storage capabilities, some methods have been introduced accordingly. In \cite{mfeng2023}, Feng \emph{et al.} applied FL to enhance the performance of Acoustic Event Classification (AEC) and improve downstream AEC classifiers by performing SSL with small negative samples locally without labels for saving computing and storage resources, and adjusting model parameters globally with labels. However, this method does not fully leverage the advantage of SSL learning without labels. In \cite{jshi2023}, instead of employing all the negative samples, Shi \emph{et al.} proposed a SSL method to automatically select a core set of most representative samples on each device and store them in a replay buffer for saving computing and storage capabilities of devices. In \cite{sweiFedCo}, Wei \emph{et al.} combined FL and MoCo to propose a FSSL named Federated learning with momentum Contrast (FedCo) for IoV, where MoCo is a special SSL method which employs dictionary to provide the reference of negative samples, which can save computing resource and storage space \cite{xchen2020MoCoV2}. However, for FedCo, vehicles uploaded local $k$ values for forming a new big dictionary in global side, which fails to ensure privacy protection and goes against the original intention of using FL. In \cite{czhang2022dual temp}, Zhang \emph{et al.} proposed the SSL method SimCo base on MoCo, and utilized dual temperatures to control sample distribution, thus SimCo not only saves computing resource and storage space but also protects privacy. To the best of our knowledge, there is no related work proposed a FSSL based on SimCo to reduce computing resource and storage space while protecting privacy in IoV. Furthermore, the training performance of FSSL in IoV may be deteriorated due to the motion blur, and there is no related work that considers motion blur in the related works that employs FSSL in IoV. Hence, we consider the motion blur and propose a FSSL based SimCo for IoV to resist motion blur, referred as to BFSSL.
	
	\subsection{DRL-based resource allocation scheme in FL}
	Some works have proposed DRL-based resource allocation schemes in IoV \cite{2}\cite{4}. In \cite{yzheng2022}, Zheng \emph{et al.} proposed a resource allocation scheme based on DRL method to minimize the duration required for accomplishing tasks while respecting latency limitations in IoV. In \cite{bhazarik2022}, Hazarika \emph{et al.} proposed a resource allocation scheme based on DRL method to allocate the power optimally in IoV. In \cite{jguo2023}, Guo \emph{et al.} proposed a resource allocation scheme based on DRL method to minimize the cost of Road Side Unit (RSU) in IoV. However, the methods mentioned above have not considered the FL framework. There are also some works that consider the FL framework to address resource allocation problems based on DRL. In \cite{yhe2023}, He \emph{et al.} combined FL and DRL method (i.e., Deep Deterministic Policy Gradient (DDPG)), to optimize both computation offloading and resource allocation. In \cite{lzhang2022}, Zhang \emph{et al.} combined FL and DRL method (i.e., DDPG), and proposed a dynamic computation offloading and resource allocation scheme to minimize the delay and energy consumption. In \cite{21}, Zhao \emph{et al.} designed a multi-objective optimization problem to assist Hierarchical Federated Learning (HFL) with Mobile Edge Computing (MEC), aiming to achieve more accurate models and reduce overhead. In \cite{22}, Zhou \emph{et al.} proposed a joint RL method based on local differential privacy. This method allows devices to add noise during local training, with the central server then designing the optimal resource allocation strategy. In \cite{23}, Zhang \emph{et al.} formulated a complex non-convex problem with the goal of minimizing the latency and energy consumption required for task execution. They proposed a multimedia task offloading and resource allocation scheme based on Federated Deep Q-Network. The aforementioned methods highlight the crucial role of DRL algorithms in addressing complex non-convex problems. However, these algorithms are primarily designed for static environments. In scenarios characterized by high mobility and rapid environmental changes, these algorithms will not offer efficient solutions.
	
	Overall, there is no work concerning the DRL-based resource allocation schemes for FSSL in IoV. Therefore, we further investigate a DRL-based resource allocation scheme to optimize the performance for the BFSSL in IoV,  referred to as DRL-BFSSL.
	
	\begin{figure}
		\centering
		\includegraphics[scale=0.35, trim=1cm 1cm 4cm 0.5cm, clip]{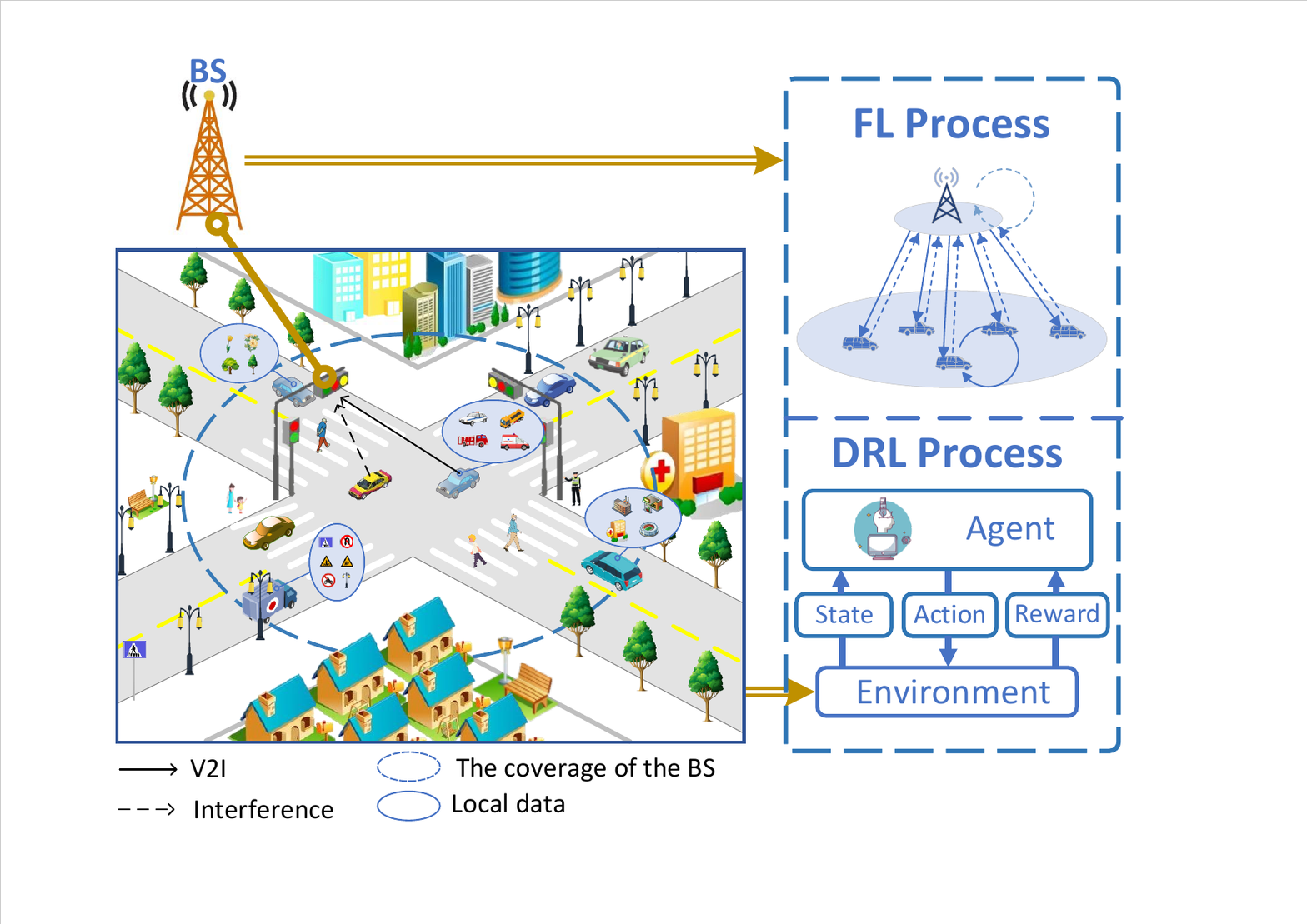}
		\caption{System scenario}
		\label{fig2}
	\end{figure}
	
	\section{System Model}
	\label{sec3}
	
	In this section, we will introduce the system scenario and models. As shown in Fig. \ref{fig2}, we consider a urban scenario of a BS deployed at the center of the intersection and several vehicles driving in the coverage of the BS. When a vehicle arrives at the intersection, it turns left, turns right and drives straightly with probabilities $\varepsilon_1$, $\varepsilon_2$, $\varepsilon_3$, and $\varepsilon_1+\varepsilon_2+\varepsilon_3=1$, and then it keeps driving straightly. Vehicle velocity follows the truncated Gaussian distribution. Each vehicle is equipped with a camera to capture images during driving through the coverage area of the BS, and the vehicles coming from different directions capture different categories of images. $N$ training vehicles are randomly selected from the vehicles before crossing the intersection to ensure that vehicles have enough time for local training. Vehicle $n$, $n\in[1,N]$, is equipped with a camera and captures $Z$ image data with velocity $v_{0}^{n}$ before it enters the coverage area of the BS.
	
	The entire process of DRL-BFSSL algorithm consists of $K^{\text{max}}$ episodes, each of which is divided into two task lines, one for BFSSL, where episode $k$ is divided into different rounds, and the other for DRL-based resource allocation, where episode $k$ is divided into different slots. 
	
\begin{figure*}[htbp]
		\centering
		\includegraphics[scale=0.5, trim=0.3cm 5cm 0.51cm 2cm, clip]{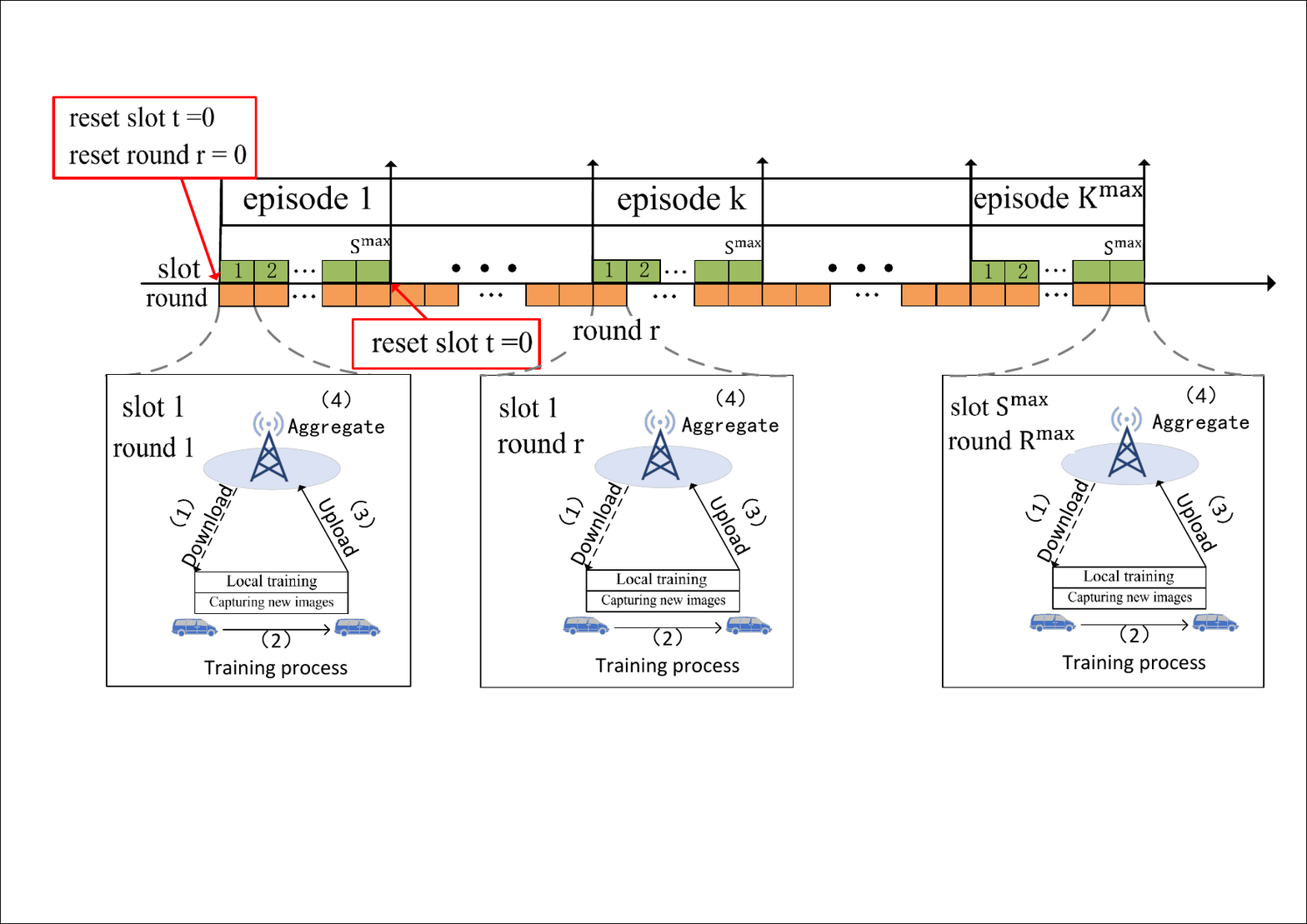}
		\caption{The process of DRL-BFSSL algorithm}
		\label{fig1111}
	\end{figure*}

	Specifically, as shown in Fig. \ref{fig1111}, within each episode $k$, $k\in[1,K^{\text{max}}]$ , there are $S^{\text{max}}$ slots. Each green cell represents a slot, and each orange cell represents a round. One slot corresponds to one round. The difference is that at the end of each episode $k$, when entering the next episode $k+1$, slot $t$ is reset to slot $0$ and start with slot $1$. In contrast, round $r$ is not reset and continues to accumulate until the whole process of DRL-BFSSL algorithm finished.
	Due to each episode containing $S^{\text{max}}$ slots, and round $r$ resets to 0 after the completion of the $K^{\text{max}}$ episodes, the relationship between slot $t$ in episode $k$ and round $r$ is as follows
	 \begin{equation}
	 	r=\left(k-1\right)S^{\text{max}}+t.
	 	\label{eq50}
	 \end{equation}
	At the beginning of the DRL-BFSSL algorithm, BS stores the parameters of the global model, while each vehicle maintains a local model with the same structure as the global model. Additionally, each vehicle keeps images captured before training, referred to as local data. Vehicles also upload their velocities to the BS, and then BS estimates their positions with the received velocities. Based on these estimates, BS quickly allocates transmission powers and CPU frequencies to vehicles. In each round of SSL, each vehicle first downloads the transmission power, CPU frequency, and the parameters of the global model from BS, and applies the parameters to its local model. The vehicles then train their local models using the local data according to the downloaded CPU frequency. After training is complete, the vehicles upload the trained parameters of their local models to BS with the assigned transmission power.
	During the transmission process, varying signals transmitted by other vehicles may cause interference, leading to data errors that prevent the BS from receiving the uploaded local models. Finally, BS aggregates the received local models to obtain a new global model based on the motion blur. The above process is repeated in different episode to get a converged global model.
	
	In the following, we will introduce the mobility model, transmission model, computing model, and data error model in slot $t$ of each episode $k$.
	
	\subsection{Mobility model}
		
	As the velocity of each vehicle $n$ follows the truncated Gaussian distribution to reflect real vehicle mobility, the velocities of different vehicles are IID. Let $v_{k,t}^{n}$ be the velocity of vehicle $n$, and $v^{\text{min}}$ and $v^{\text{max}}$ be the minimum and maximum velocity of vehicles, respectively. Thus the probability density function of $v_{k,t}^{n}$ is calculated as \cite{zyu2021}
	\begin{equation}
			\vspace{-0.25cm}
			f\left(v_{k,t}^{n}\right) = \left\{ \begin{aligned}
				\frac{{{e^{ - \frac{1}{{2{\sigma ^2}}}{{({v_{k,t}^{n}} - \mu )}^2}}}}}{{\sqrt {2\pi {\sigma ^2}} [\text{erf}(\frac{{{v^{\max }} - \mu }}{{\sigma \sqrt 2 }}) - \text{erf}(\frac{{{v^{\min }} - \mu }}{{\sigma \sqrt 2 }})]}}\\
				{v^{\text{min}}} \le {v_{k,t}^{n}} \le {v^{\text{max}}}\\
				0 \qquad \qquad \qquad \qquad \quad \text{otherwise}
			\end{aligned} \right.,
			\label{eq1}
	\end{equation}where $\text{erf}(\mu, \sigma^2)$ is the Gaussian error function of velocity $v_{k,t}^{n}$ with mean $\mu$ and variance $\sigma^2$. 
		
	\subsection{Transmission model}
	During the transmission process, the energy consumption of vehicle $n$ can be calculated as
	\begin{equation}
	E_{k,t}^{{\text{trans}},n}=p_{k,t}^{n} T_{k,t}^{{\text{trans}},n},
		\label{eq17}
	\end{equation}where $p_{k,t}^{n}$ is the transmission power of vehicle $n$, $T_{k,t}^{{\text{trans}},n}$ is the transmission delay required by vehicle $n$ to transmit local model. We consider that each vehicle transmits a local model with the the same size $Z$, thus $T_{k,t}^{{\text{trans}},n}$ can be calculated as
	\begin{equation}
		T_{k,t}^{{\text{trans}},n}=\frac{Z}{c_{k,t}^{n}},
			\label{eq15}
	\end{equation}
	where $c_{k,t}^{n}$ is the transmission rate of vehicle $n$, which can be calculated according to Shannon's theorem \cite{sluo2020}, i.e., 
	\begin{equation}
	c_{k,t}^{n}=\beta_{k,t}^{n}B^U\text{ln}\left(1+\frac{p_{k,t}^{n}h_{k,t}^{n}}{I_{k,t}^{n}+B^UN_0}\right),
			\label{eq14}
	\end{equation}
	where $\beta_{k,t}^{n}$ is the proportion of the bandwidth allocated to vehicle $n$, $B^U$ is the uplink bandwidth, $I_{k,t}^{n}$ is interference and $N_0$ is the noise power. $h_{k,t}^{n}$ is the channel power gain between vehicle $n$ and the BS, which is calculated as \cite{lliang2017}:
	\begin{equation}
	h_{k,t}^{n}=g_{n}\varrho_{n}\mathcal{A}\left(L_{k,t}^{n}\right)^{-\iota},
	\label{eq16}
	\end{equation}
	where $g_{n}$ represents the small-scale fast fading power, which follows an exponential distribution with unit mean. $\varrho_{n}$ represents shadow fading with a standard deviation of $\xi$. $\mathcal{A}$ is the path loss constant, $L_{k,t}^{n}$ is the distance between vehicle $n$ and BS, and $\iota$ is the decay exponent \cite{ReconfigurableHolographic}.
		
	\subsection{Computing model}
	During the local training, the energy consumption of vehicle $n$ to execute each iteration of local training is calculated as
	\begin{equation}
	E_{k,t}^{{\text{comp}},n}=p_{k,t}^{{\text{comp}},n} T_{k,t}^{{\text{comp}},n},
	\label{eq25}
	\end{equation} where $T_{k,t}^{{\text{comp}},n}$ is the computing delay of vehicle $n$ to execute one iteration of local training, and $p_{k,t}^{{\text{comp}},n}$ is the CPU computing power of vehicle $n$, that can be calculated by Dynamic Voltage Frequency Scaling (DVFS) approach \cite{hxiao2021}, i.e.,
	\begin{equation}
	p_{k,t}^{{\text{comp}},n}=\kappa\left(f_{k,t}^{n}\right)^3, 
	\label{eq19}
	\end{equation}
	where $\kappa$ is the effective switched capacitance, which depends on the chip architecture, and $f_{k,t}^{n}$ is the CPU frequency allocated to vehicle $n$ for computation. Let $f^{\text{min}}$ and $f^{\text{max}}$ be the minimum and maximum CPU computation frequency, thus $f_{k,t}^{n}\in[f^{\text{min}},\ f^{\text{max}}]$.
		
	Let $r_{\text{cyc}}$ be the number of CPU cycles for processing the unit size and $D_{\text{data}}$ be the local data size. $|D_{\text{data}}|r_{\text{cyc}}$ represents the required CPU cycles in local training. Hence, the computing delay $T_{k,t}^{{\text{comp}},n}$ is calculated as
	\begin{equation}
	T_{k,t}^{{\text{comp}},n}=\frac{\left|D_{\text{data}}\right|r_{\text{cyc}}}{f_{k,t}^{n}}.
		\label{eq18}
	\end{equation}
	
	Substituting Eqs. \eqref{eq19} and \eqref{eq18} into Eq. \eqref{eq25} , we can obtain
	\begin{equation}
		E_{k,t}^{{\text{comp}},n}=\kappa\left(f_{k,t}^{n}\right)^2\left|D_{\text{data}}\right|r_{\text{cyc}}.
		\label{eq100}
	\end{equation}

	To evaluate the utility of CPU frequency, we use the inverse of energy consumption as the criterion for the utility function. The utility $U$
	can be expressed as \cite{WWH}
		\begin{equation}
		U=\frac{1}{E_{k,t}^{{\text{comp}},n}}.
		\label{eq101}
	\end{equation}
	\subsection{Data error model}
	
	In data transmission, both the variations signals transmitted by other vehicles may cause interference that incurs an error probability for the data of the local model received by the BS. Cyclic Redundancy Check (CRC) mechanism is employed here to describe error for the data of local model received by the BS \cite{mchen2020}. According to the CRC mechanism, the data error probability $\epsilon_{k,t}^{n}$ for vehicle $n$ in slot $t$ of episode $k$ is calculated as\cite{yxi2011}
	\begin{equation}
		\epsilon_{k,t}^{n}\left(p_{k,t}^{n}\right)=1-\text{exp}\left[-\frac{m\left(I_{k,t}^{n}+B^UN_0\right)}{p_{k,t}^{n}h_{k,t}^{n}}\right],
		\label{eq2}
	\end{equation}
	where $m$ is a waterfall threshold.
	Let $C_{k,t}^{n}(p_{k,t}^{n})$ indicates if the BS successfully received the uploaded model with respect to $p_{k,t}^{n}$. Thus we have
	
	\begin{equation}
		\vspace{-0.00cm}
		C_{k,t}^{n}(p_{k,t}^{n}) = \begin{cases}
			1, & \text{with probability } (1 - \epsilon_{k,t}^{n}) \\
			0, & \text{with probability } \epsilon_{k,t}^{n}
		\end{cases}.
		\label{eq333}
	\end{equation}

	\section{Optimization Problem}
	\label{sec4}
	
	The duration of one round is donated as $T$, which is divided into two phases: local training phase and uploading phase. 
	
	In the local training phase, higher CPU frequency allows for more iterations, which can enhance model performance but also increase energy consumption. In the uploading phase, vehicles upload the trained local model to BS.  Higher transmission power reduces the upload time but increases energy consumption. To save resources, we need to optimize the allocation of vehicle CPU frequencies and transmission power to minimize the overall energy consumption of all vehicles \cite{HybridNearFar}.
	
	However, if the CPU frequency is too low, the time for a single local training iteration increases, resulting in fewer local iterations. Although this reduces energy consumption, it does not ensure the model's performance. Therefore, we also need to minimize the time for each training to ensure sufficient local iterations \cite{Delayconstrained}\cite{SemanticAware}.
	
	To deal with these two problems, our target is adjusted to minimize the total energy consumption of all vehicles and the max delay including model training and model transmission \cite{1}\cite{5}. The energy and delay for system performance can be calculated as:
		\begin{equation}
	E_{k,t}^{{\text{total}},n}=E_{k,t}^{{\text{trans}},n}+\left\lfloor N_{k,t}^{n}\right\rfloor E_{k,t}^{{\text{comp}},n},
		\label{eq222}
	\end{equation}
	\begin{equation}
		T_{k,t}^{{\text{total}},n}=T_{k,t}^{{\text{trans}},n}+T_{k,t}^{{\text{comp}},n},
		\label{eq233}
	\end{equation}
	where$\left\lfloor N_{k,t}^{n}\right\rfloor$ is the number of iterations for local training, $\left\lfloor\cdot\right\rfloor$ represents the rounding down function. Since the local iterations should be a integer, the integral number of iterations $\left\lfloor N_{k,t}^{n}\right\rfloor$ can be calculated as
	\begin{equation}
		\left\lfloor  N_{k,t}^{n}\right\rfloor=\left\lfloor\frac{T-t_{\max}^{\text{Trans}}}{T_{k,t}^{{\text{comp}},n}}\right\rfloor,
		\label{eq20}
	\end{equation} 
	where $t_{\max}^{\text{Trans}}$ denotes the maximum allowable transmission delay. For a fixed $T$, $\left(T-t_{\max}^{\text{Trans}}\right)$ is the time used for local training. For simplicity, we use $\mathcal{N}_{k,t}^{n}$ to approximate the value $	\left\lfloor  N_{k,t}^{n}\right\rfloor$, where
	
	\begin{equation}
		\mathcal{N}_{k,t}^{n}=N_{k,t}^{n}-1.
		\label{eq111}
	\end{equation}

	Thus Eq. \eqref{eq222} can be expressed as
	\begin{equation}
		\begin{split}
			E_{k,t}^{{\text{total}},n}=E_{k,t}^{{\text{trans}},n}&+\mathcal{N}_{k,t}^{n} E_{k,t}^{{\text{comp}},n}, \\&
			\left\lfloor  N_{k,t}^{n}\right\rfloor-1\le {\mathcal{N}_{k,t}^{n}}\le	\left\lfloor  N_{k,t}^{n}\right\rfloor.
			\label{eq22}
		\end{split}
	\end{equation}
	
	Let $\boldsymbol{\beta}=\left\{\beta_{k,t}^{1},\beta_{k,t}^{2},\cdots,\beta_{k,t}^{n},\cdots,\beta_{k,t}^{N}\right\}$,
	
	$\boldsymbol{p}=\left\{{p_{k,t}^{1},p_{k,t}^{2},\cdots,p_{k,t}^{n},\cdots,p_{k,t}^{N}}\right\}$, and $\boldsymbol{f}=\left\{f_{k,t}^{1},f_{k,t}^{2},\cdots,f_{k,t}^{n},\cdots,f_{k,t}^{N}\right\}$ represent the sets of the allocated bandwidth ratios, transmission powers, and CPU frequencies, respectively. The optimization problem can be formulated as
	
	\begin{equation}
		\underset{\boldsymbol{\beta}, \boldsymbol{p}, \boldsymbol{f}}{\min} \mathcal{C} : \lambda_1\sum_{n=1}^{N}{	E_{k,t}^{{\text{total}},n}+\lambda_2\max{	T_{k,t}^{{\text{total}},n}}}
		\label{eq24}\\
	\end{equation}
	\hspace{5em}\text{s.t.}
	\begin{subequations}
		\begin{equation}
			0 < \beta_{k,t}^{n} < 1\tag{\ref{eq24}{a}} \label{eq24a}
		\end{equation}
		\begin{equation}
			\sum_{n=1}^{N} \beta_{k,t}^{n} \le 1 \tag{\ref{eq24}{b}} \label{eq24b}
		\end{equation}
		\begin{equation}
			\epsilon_{k,t}^{n} \le \epsilon_\tau \tag{\ref{eq24}{c}} \label{eq24c}
		\end{equation}
		\begin{equation}
			p^{\text{min}} \le p_{k,t}^{n} \le p^{\text{max}}\tag{\ref{eq24}{d}} \label{eq24d}
		\end{equation}
		\begin{equation}
			f^{\text{min}} \le f_{k,t}^{n} \le f^{\text{max}}\tag{\ref{eq24}{e}} \label{eq24e}
		\end{equation}
	\end{subequations}
	where $\lambda_1$ and $\lambda_2$ represent the weights assigned to the energy consumption and delay, respectively, thus $\lambda_1,\ \lambda_2\in[0,1]$, and $\lambda_1+\lambda_2=1$. Eqs. \eqref{eq24a} and \eqref{eq24b} show that the allocated bandwidth for each vehicle $n$ maintains a reasonable range. Eq. \eqref{eq24c} ensures that the data error rate of vehicle $n$ is less than a specified value $\epsilon_\tau$. Eq. \eqref{eq24d} limits the minimum and maximum transmission power for each vehicle $n$, i.e., $p^{\text{min}}$ and $p^{\text{max}}$. Eq. \eqref{eq24e} limits the minimum and maximum CPU frequency for vehicle $n$, i.e., $f^{\text{min}}$ and $f^{\text{max}}$.
	
	As can be seen from the above optimization problem, the restrictions Eqs. \eqref{eq24a} - \eqref{eq24e} increase the complexity of calculation and the difficulty of problem solving. Next we will simplify the restrictions.
	
	Substituting Eqs. \eqref{eq222}, \eqref{eq233} and \eqref{eq111} into Eq. \eqref{eq24}, the optimization problem can be written as
	\begin{equation}
		\begin{aligned}
			&\mathcal{C} = \lambda_1 \sum_{n=1}^{N} E_{k,t}^{{\text{total}},n} + \lambda_2\max{	T_{k,t}^{{\text{total}},n}} \\
			&=\sum_{n=1}^{N} \Big[ \frac{\lambda_1 p_{k,t}^{n} Z}{\beta_{k,t}^{n} B^U \ln\left(1 + \frac{p_{k,t}^{n}h_{k,t}^{n}}{I_{k,t}^{n} + B^UN_0}\right)} \\&+ \lambda_1 \left( T - t_{\max}^{\text{Trans}} \right) \kappa \left(f_{k,t}^{n}\right)^3 - \lambda_1 \kappa \left(f_{k,t}^{n}\right)^2 |D_{\text{data}}|r_{\text{cyc}}\Big] \\
			&+\max\Big[\frac{\lambda_2 Z}{\beta_{k,t}^{n} B^U \ln\left(1 + \frac{p_{k,t}^{n}h_{k,t}^{n}}{I_{k,t}^{n} + B^UN_0}\right)}+ \frac{\lambda_2 |D_{\text{data}}|r_{\text{cyc}}}{f_{k,t}^{n}}\Big].
		\end{aligned}
		\label{eq27}
	\end{equation}
	
	As shown in Eq. \eqref{eq27}, the optimization problem involves multiple variables. To simplify the solving process, we introduce the following notations
	\begin{equation}
		A=\frac{\lambda_1p_{k,t}^{n}Z}{B^U \text{ln}\left(1+\frac{p_{k,t}^{n}h_{k,t}^{n}}{I_{k,t}^{n}+B^UN_0}\right)},
		\label{eq28}
	\end{equation}
	\begin{equation}
		B=\lambda_1\kappa\left(T-t_{\max}^{\text{Trans}}\right),
		\label{eq29}
	\end{equation}
	\begin{equation}
		C=\lambda_1\kappa\left|D_{\text{data}}\right|r_{\text{cyc}},
		\label{eq30}
	\end{equation}
	\begin{equation}
		E=\frac{\lambda_2Z}{B^U\text{ln}\left(1+\frac{p_{k,t}^{n}h_{k,t}^{n}}{I_{k,t}^{n}+B^UN_0}\right)},
		\label{eq31}
	\end{equation}
	\begin{equation}
		F=\lambda_2\left|D_{\text{data}}\right|r_{\text{cyc}},
		\label{eq32}
	\end{equation}
	where $A$, $B$, $C$, $E$, and $F$ depend on the parameter $p_{k,t}^{n}$ and the constant parameters related to the vehicle's own configuration. Thus the optimization problem $\mathcal{C}$ is written as 
	
	\begin{equation}
		\begin{split}
			\mathcal{C} = & \sum_{n=1}^{N}\left[\frac{A}{\beta_{k,t}^{n}}+B\left({f_{k,t}^{n}}\right)^3-C\left({f_{k,t}^{n}}\right)^2\right] \\
			&+ \max{\left(\frac{E}{\beta_{k,t}^{n}}+\frac{F}{f_{k,t}^{n}}\right)}.
		\end{split}
		\label{eq33}
	\end{equation}
	
	Then we utilize the KKT conditions to reduce the restrictions \eqref{eq24a} and \eqref{eq24b}, while the detailed procedures is given in Appendix A. Thus we can obtain the relationship between $\beta_{k,t}^{n}$ and the introduced notations, i.e.,
	
	\begin{equation}
		\beta_{k,t}^{n}=\frac{\left[A+\frac{3B\left({f_{k,t}^{n}}\right)^4-2C\left({f_{k,t}^{n}}\right)^3}{F}E\right]^\frac{1}{2}}{\sum_{n=1}^{N}\left[A+\frac{3B\left({f_{k,t}^{n}}\right)^4-2C\left({f_{k,t}^{n}}\right)^3}{F}E\right]^\frac{1}{2}}.
		\label{eq35}
	\end{equation}
	
	According to the Eq. \eqref{eq2} and restriction \eqref{eq24c}, we can establish the relationship between $\epsilon_{\tau}$ and $p_{k,t}^{n}$, thus Eq. \eqref{eq2} can be written as 
	\begin{equation}
		p_{k,t}^{n}\geq\frac{-m\left(I_{k,t}^{n}+B^UN_0\right)}{h_{k,t}^{n}\text{ln}\left(1-\epsilon_\tau\right)}.
		\label{eq26}
	\end{equation} 
	We define the right-hand side of Eq. \eqref{eq26} as $P^\tau$. Thus, the optimization problem $\mathcal{C}$ can be formulated as:
	\begin{equation}
		\begin{split}
			\underset{\boldsymbol{p}, \boldsymbol{f}}{\min} \mathcal{C} : &\sum_{n=1}^{N} \Bigg\{\frac{A{\sum_{n=1}^{N}\left[A + \frac{3B\left({f_{k,t}^{n}}\right)^4 - 2C\left({f_{k,t}^{n}}\right)^3}{F}E\right]^\frac{1}{2}}}{\left[A + \frac{3B\left({f_{k,t}^{n}}\right)^4 - 2C\left({f_{k,t}^{n}}\right)^3}{F}E\right]^\frac{1}{2}} \\
			&+ B\left({f_{k,t}^{n}}\right)^3 - C\left({f_{k,t}^{n}}\right)^2\Bigg\}
			\\
			&	+ \max \Bigg\{\frac{E{\sum_{n=1}^{N}\left[A + \frac{3B\left({f_{k,t}^{n}}\right)^4 - 2C\left({f_{k,t}^{n}}\right)^3}{F}E\right]^\frac{1}{2}}}{\left[A + \frac{3B\left({f_{k,t}^{n}}\right)^4 - 2C\left({f_{k,t}^{n}}\right)^3}{F}E\right]^\frac{1}{2}}\\& + \frac{F}{f_{k,t}^{n}}\Bigg\}\label{eq37}\\
		\end{split}
	\end{equation}
	\text{s.t.}
	\begin{subequations}
		\begin{equation}
			\max{(p^{\text{min}},P^\tau)}\le p_{k,t}^{n}\le p^{\text{max}} \tag{\ref{eq37}{a}} \label{eq37a}
		\end{equation}
		\begin{equation}
			f^{\text{min}}\le f_{k,t}^{n}\le f^{\text{max}}\tag{\ref{eq37}{b}}. \label{eq37b}
		\end{equation}
	\end{subequations}
	
	\section{DRL-BFSSL Algorithm}
	\label{sec5}
	The optimization problem $ \mathcal{C}$ is a NonLinear Programming (NLP) problem. As the number of the vehicles increases, it is hard to address rapidly with the traditional optimization algorithms. Thus, to achieve the optimal solution, we will consider a DRL offloading process. As we know, the SAC algorithm can deal with continuous variables in Non-convex function, and keep robustness and adaptability \cite{DistributedDeep,AwareResource}. Moreover, SAC is an off-policy DRL algorithm designed to enhance both the exploration of the policy and the system stability of the learning process. It achieves this by leveraging the maximum entropy framework, which encourages exploration by maximizing the entropy of the selected policy. This approach helps prevent premature convergence to suboptimal solutions. Additionally, SAC employs double Q-networks to mitigate the overestimation bias commonly encountered in value function approximation, further improving the model stability and reliability of the learning process \cite{CooperativeEdge,qkw}. Thus, we adopt SAC algorithm to solve the optimization problem $\mathcal{C}$. In SAC algorithm, it consists of five Deep Neural Networks (DNNs), i.e., one actor network $\varsigma$, two critic networks $\xi_1$ and $\xi_2$, and two target critic networks $\varpi_1$ and $\varpi_2$. The two critic networks work to enhance network performance by minimizing positive bias. The two target networks have the same structure as the critic networks, aiming to improve training speed and stability. By utilizing reward function and Stochastic Gradient Descent (SGD) algorithm, SAC allows us to address the problem in continuous space.
	
	In episode $k$, for each slot $t$, the decision-making includes state, action and reward. Next, we will introduce the process of obtaining these components.
	
	\subsubsection{State}
	At the beginning of episode $k$, slot $t$, the BS estimates the distance $L_{k,t}^{n}$, the unit is meters, between vehicle $n$ and BS according to the position and uploaded velocity from vehicle $n$. Then, BS can acquire fading information $\mathcal{J}_{k,t}^{n}$ between itself and vehicle $n$ which includes path loss $\mathcal{A}_{k,t}^{n}$ and shadow fading $\mathcal{E}_{k,t}^{n}$. In mobile communication systems, path loss $\mathcal{J}_{k,t}^{n}$ can be calculated as 
	\begin{equation}
		\mathcal{J}_{k,t}^{n}=10^{\frac{\mathcal{A}_{k,t}^{n}}{10}}10^{\frac{\mathcal{E}_{k,t}^{n}}{10}},
		\label{eq38}
	\end{equation}
	where $\mathcal{E}_{k,t}^{n}$ follows a normal distribution with unit mean and a standard deviation, and in modeling wireless signal propagation, $\mathcal{A}_{k,t}^{n}$ can be calculated by\cite{lliang2017}
	\begin{equation}
		\mathcal{A}_{k,t}^{n}=128.1+37.6\lg{\frac{L_{k,t}^{n}}{1000}},
		\label{eq36}
	\end{equation}where 128.1 is a constant term, while 37.6 is the path loss exponent.
	
	We consider the fading information from all training vehicles, denoted as ${\mathcal{J}}_{k,t}=\left\{\mathcal{J}_{k,t}^{1},\mathcal{J}_{k,t}^{2}, \mathcal{J}_{k,t}^{n},\cdots, \mathcal{J}_{k,t}^{N}\right\}$ and the velocity $\mathcal{V}_{k,t}=\left\{v_{k,t}^{1},v_{k,t}^{2}, v_{k,t}^{n},\cdots, v_{k,t}^{N}\right\}$ according to mobility model in Eq. \eqref{eq1}  as the state's elements. In summary, for slot $t$, the state can be expressed as
	\begin{equation}
		s_{k,t}=\left[{\mathcal{J}}_{k,t},\mathcal{V}_{k,t}\right].
		\label{eq40}
	\end{equation}
	
	\subsubsection{Action}
	In accordance with optimization problem $\mathcal{C}$, two variables, transmission power $p_{k,t}^{n}$ and CPU frequency $f_{k,t}^{n}$ need to be determined. Note that $\mathcal{P}_{k,t}=\left\{p_{k,t}^{1},p_{k,t}^{2},\cdots,p_{k,t}^{n},\cdots,p_{k,t}^{N}\right\}$ and $\mathcal{F}_{k,t}=\left\{f_{k,t}^{1},f_{k,t}^{2},\cdots,f_{k,t}^{n},\cdots,f_{k,t}^{N}\right\}$. The action can be expressed as
	\begin{equation}
		a_{k,t}=\left[\mathcal{P}_{k,t},\mathcal{F}_{k,t}\right].
		\label{eq41}
	\end{equation}
	
	\subsubsection{Reward}
	
	Firstly, according to Eq. \eqref{eq37a}, due to the constraints of $P^\tau$ on $p^{\text{min}}$, we define
	\begin{equation}
		P^\ast=\max{\left(p^{\text{min}}, P^\tau\right)},
		\label{eq42}
	\end{equation}where $P^\ast$ is the lower bound of the range for transmission power modified by $P^\tau$.
	To ensure that the range of transmission power $p_{k,t}^{n}$ of vehicle $n$ in slot $t$ is as large as possible, the difference between $p^{\text{min}}$ and $P^\ast$ should be as small as possible. $P^\ast - p^{\text{min}} = 0$ indicates that $P^\tau$ does not effect the transmission power range, maintaining the maximum range, whereas $P^\ast - p^{\text{min}} = P^\tau - p^{\text{min}}$ implies that $P^\tau$ will serve as the lower bound, and narrow the range of transmission power. While minimizing the optimization problem and the difference between $p^{\text{min}}$ and $P^\ast$, we still aim to increase the number of local iterations to enhance the performance of the global model.
	Thus, reword can be defined as
	\begin{equation}
		r_{k,t}=-\left[\mathcal{C}+\vartheta_1(P^\ast-p^{\text{min}})\right]+\vartheta_2\sum_{n=1}^{N}\mathcal{N}_{k,t}^n,
		\label{eq43}
	\end{equation}and $\vartheta_1$ and $\vartheta_2$ denote the penalty coefficient. 
	
	\subsubsection{SAC-based solution}
	
	In this section, we will introduce our proposed DRL-BFSSL algorithm. 
	
	\textbf{\emph{Step 1, initialization}}: BS stores the parameter of a global model and five networks of SAC algorithm, while each vehicle $n$ stores $\mathcal{Z}$ local image data and the velocity $v_{0}^{n}$ mapping the image data. Meanwhile, vehicle $n$ also stores a model with the same structure as global model, referred to as local model. At the beginning of the algorithm, BS randomly initializes the parameters of the global model $\theta_0^{{\text{global}}}$, along with the actor network $\varsigma$, two critic networks $\xi_1$ and $\xi_2$, and also assign the parameters of two critic networks to the two target networks $\varpi_1$ and $\varpi_2$. Vehicle $n$ uploads the velocity $v_1^n$ for slot $1$. 
	
	\textbf{\emph{Step 2, local training}}: At the beginning of episode $k$, slot $t$, setting the value of round $r$ according to Eq. \eqref{eq50} firstly.
	
	BS stores the parameters of a global model $\theta_{r-1}^{{\text{global}}}$ which is aggregated at the end of previous round $r-1$ and estimates the positions according to the uploaded velocities $ \left\{v_{k,t}^{1},v_{k,t}^{2},\ldots v_{k,t}^{n}\ldots v_{k,t}^{N}\right\}$. Then BS can get state $s_{k,t}$ according to Eq. \eqref{eq40}. Input $s_{k,t}$ into actor network and output $a_{k,t}$, including transmission powers and CPU frequencies for vehicles. 
	Then, vehicle $n$ downloads the parameters of the global model $\theta_{r-1}^{{\text{global}}}$, transmission power $p_{k,t}^{n}$ , CPU frequency $f_{k,t}^{n}$ from the BS, and sets the parameter of global model to local model $\theta_{r}^{{\text{local},n}}$. If in the first round, the parameters of global model comes from $\theta_0^{{\text{global}}}$. If in the episode $k$, slot $1$, the parameters of SAC networks come from the output of episode $k-1$, slot $S^{\text{max}}$.
	
	\begin{figure}
		\centering
		\includegraphics[scale=0.31, trim=2cm 4cm 1cm 3cm, clip]{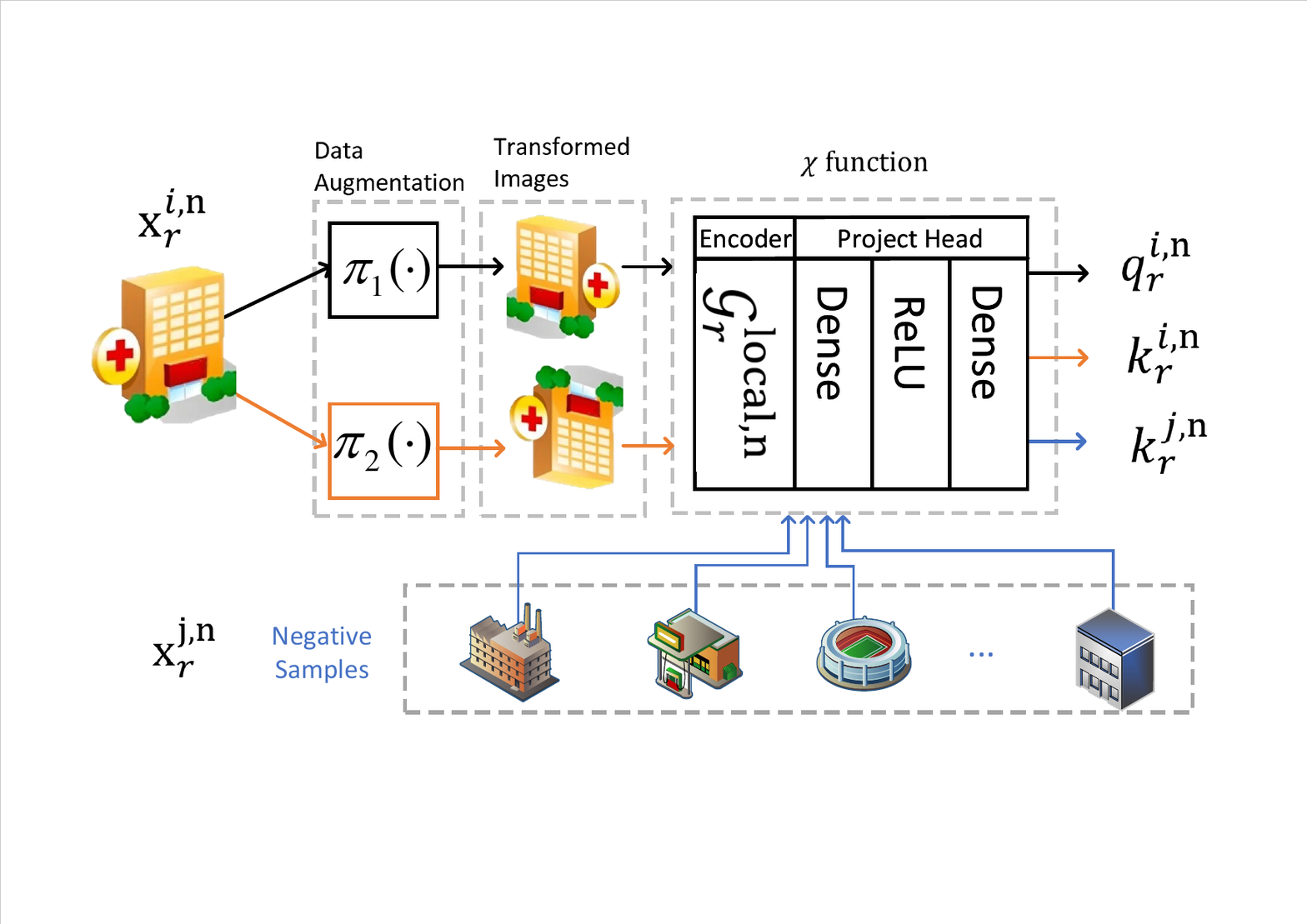}
		\caption{Encoding process}
		\label{fig3}
	\end{figure}
	
	Next, the maximum number of iterations $\mathcal{N}_{k,t}^{n}$ of local training for vehicle $n$ is calculated according to Eq. \eqref{eq20} and Eq. \eqref{eq111} based on the downloaded CPU frequency $f_{k,t}^{n}$. Then each training vehicle retrains the local model for $\mathcal{N}_{k,t}^{n}$ iterations. Specifically, for each iteration $\psi_{k,t}^n\in[1,\mathcal{N}_{k,t}^{n}]$ of local training, each vehicle $n$ encodes the images of the local data, where the encoding process is illustrated in Fig. \ref{fig3}. For each image $x_{r}^{i,n}\in\left\{x_{r}^{1,n},\ x_{r}^{2,n},\ \ldots x_{r}^{i,n},\ldots x_{r}^{\mathcal{Z},n}\right\}$ in vehicle $n$ undergoes two different data augmentation methods $\pi_1(\cdot)$ and $\pi_2(\cdot)$. $\pi_1(\cdot)$ performs a horizontal flip on the image with a 50\% probability, followed by converting the image to grayscale with a 20\% probability. $\pi_2(\cdot)$ randomly alters the image's brightness, contrast, saturation, and hue with an 80\% probability, followed by converting the image to grayscale with a 40\% probability. It is also noted that $\pi_1(\cdot)$ and $\pi_2(\cdot)$ share the same original image when they process the image in different way \cite{gxy}. We treat the remaining images, donated as  $x_{r}^{j,n},j\in\left[1,\mathcal{Z}\right]\ \text{and}\ j\neq i$, as negative samples. Then, these two augmented images $\pi_1(x_{r}^{i,n})$ and $\pi_2(x_{r}^{i,n})$ along with $x_{r}^{j,n}$ are input into an $\chi$ function, which composed of an encoder with function $\mathcal{G}_{r}^{\text{local},n}$, and a project head including two dense neutral networks and a ReLU function. The $\chi$ function outputs anchor sample $q_{r}^{i,{n}}$, positive sample $k_{{r}}^{i,{n}}$, and encoded negative samples $k_{{r}}^{j,{n}}$, respectively, i.e.,
	
	\begin{equation}
		q_{r}^{i,n}=\chi\left\{\left[\pi_1\left(x_{r}^{i,n}\right)\right]\right\},\ i\in\left[1,\mathcal{Z}\right],
		\label{eq5}
	\end{equation}
	\begin{equation}
		k_{r}^{i,n}=\chi\left\{\left[\pi_2\left(x_{r}^{i,n}\right)\right]\right\},\ i\in\left[1,\mathcal{Z}\right],
		\label{eq6}
	\end{equation}
	\begin{equation}
		k_{r}^{j,n}=\chi\left\{\left(x_{r}^{j,n}\right)\right\},\ j\in\left[1,\mathcal{Z}\right],\ \text{and}\ j\neq i.
		\label{eq77}
	\end{equation}
	
	The dual temperature loss of the $i$-th image of anchor sample of the vehicle $n$ in round $r$ can be transformed into \cite{czhang2022dual temp}
	
	\begin{equation}
		\begin{split}
			\mathcal{L}_{q_{{r}}^{i,n}}^{\text{DT}}\left(\theta_{{r}}^{\text{\text{local}},n},\ q_{{r}}^{i,n},\ k_{{r}}^{i,n},\  k_{{r}}^{j,n}\right)=-\text{sg}\left[\frac{{{(W}_\beta)}_{{r}}^{i,n}}{{{(W}_\alpha)}_{{r}}^{i,n}}\right]\\    \times\text{log}\frac{\exp{\left(\frac{q_{{r}}^{i,n}\cdot k_{{r}}^{i,n}}{\tau_\alpha}\right)}}{\exp{\left(\frac{q_{{r}}^{i,n}\cdot k_{{r}}^{i,n}}{\tau_\alpha}\right)}+\sum_{j=1}^{\mathcal{K}}\exp{\left(\frac{q_{{r}}^{i,n}\cdot k_{{r}}^{j,n}}{\tau_\alpha}\right)}},
		\end{split}
		\label{eq7}
	\end{equation}
	where $\text{sg}[\cdot]$ indicates the stop gradient, and
	\begin{equation}
		\begin{split}
			{{(W}_\beta)}&_{{r}}^{i,n}=1-
			\\&\frac{\exp{\left(\frac{q_{{r}}^{i,n}\cdot _{{r}}^{i,n}}{\tau_\beta}\right)}}{\exp{\left(\frac{q_{{r}}^{i,n}\cdot k_{{r}}^{i,n}}{\tau_\beta}\right)}+\sum_{j=1}^{\mathcal{K}}\exp{\left(\frac{q_{{r}}^{i,n}\cdot k_{{r}}^{j,n}}{\tau_\beta}\right)}},
		\end{split}
		\label{eq8}
	\end{equation}
	\begin{equation}
		\begin{split}
			{{(W}_\alpha)}&_{{r}}^{i,n}=1-
			\\&\frac{\exp{\left(\frac{q_{{r}}^{i,n}\cdot k_{{r}}^{i,n}}{\tau_\alpha}\right)}}{\exp{\left(\frac{q_{{r}}^{i,n}\cdot k_{{r}}^{i,n}}{\tau_\alpha}\right)}+\sum_{j=1}^{\mathcal{K}}\exp{\left(\frac{q_{{r}}^{i,n}\cdot k_{{r}}^{j,n}}{\tau_\alpha}\right)}},
		\end{split}
		\label{eq9}
	\end{equation}where $\tau_\alpha$ and $\tau_\beta$ are temperature hyper-parameters \cite{khou2020}, which that control the shape of the samples distribution, and $\mathcal{K}$ is the number of negative samples. 
	
	Then, for each image, the aim is to minimize the loss function and get ideal optimize local mode $\hat{\theta}_{{r}}^{{\text{local}},n}$, i.e.,
	
	\begin{equation}
		\begin{aligned}
			\hat{\theta}_{{r}}^{{\text{local}},n} &= \underset{\theta_{{r}}^{\text{local},n}}{\operatorname{argmin}} \frac{1}{\mathcal{Z}} \sum_{i=1}^{\mathcal{Z}} \mathcal{L}_{q_{{r}}^{i,n}}^{\text{DT}} \big( \theta_{{r}}^{{\text{local}},n}, q_{{r}}^{i,n}, \\
			&\hspace{4cm} k_{{r}}^{i,n}, k_{{r}}^{j,n} \big),
		\end{aligned}
	\end{equation}where $\theta_{{r}}^{{\text{local},n}}$ represents the parameters of local model of vehicle $n$ in round $r$. However, $\hat{\theta}_{{r}}^{{\text{local}},n}$ is an ideal value, and we use SGD to gradually bring $\theta_{{r}}^{{\text{local},n}}$ closer to $\hat{\theta}_{{r}}^{{\text{local}},n}$. The updating process can be described as follows:
	\begin{align}
		\begin{split}
			\theta_{{r}}^{\text{local},n}\gets\ &\theta_{{r}}^{\text{local},n}
			\\&-\eta^{r}\nabla\mathcal{L}_{q_{{r}}^{i,n}}^{\text{DT}}\left(\theta_{{r}}^{\text{local},n},\ q_{{r}}^{i,n},\ k_{{r}}^{i,n},\  k_{{r}}^{j,n}\right),
			\label{eq12}
		\end{split}
	\end{align}where $\eta^{r}$ represents the learning rate for round $r$, and $\nabla$ is the SGD algorithm. Repeat the local training $\mathcal{N}_{k,t}^{n}$ iterations and output the final local model $\theta_{{r}}^{\text{local},n}$.
	
	Noting that during the local training process, the vehicle simultaneously captures $\mathcal{Z}$ new images $x_{r+1}^{i,n}$, $i\in\left[1,\mathcal{Z}\right]$ with the camera for the next round of local training. When the iterations reach to $\mathcal{N}_{k,t}^{n}$, the local training is finished, and outputs the final local model $\theta_{{r}}^{\text{local},n}$.
	
	\textbf{\emph{Step 3, Upload model}}: $N$ training vehicles employ downloaded transmission power to upload the trained local models $ \left\{\theta_{{r}}^{{\text{local},1}},\theta_{{r}}^{{\text{local},2}},\ldots \theta_{{r}}^{{\text{local},n}}\ldots \theta_{{r}}^{{\text{local},N}}\right\}$ and the velocities $ \left\{v_{k,t+1}^{1},v_{k,t+1}^{2},\ldots v_{k,t+1}^{n}\ldots v_{k,t+1}^{N}\right\}$ for slot $t+1$ based on the mobility model. Specifically, each vehicle will upload the local model $\theta_{{r}}^{{\text{local},n}}$ and $v_{k,t}^{n}$ with transmission power $p_{k,t}^{n}$ when local training is finished. 
	
	\textbf{\emph{Step 4, Aggregation and Update}}: Firstly, $C_{k,t}^{n}(p_{k,t}^{n})$ can be calculated by Eq. \eqref{eq2} and Eq. \eqref{eq333} based on the transmission power $p_{k,t}^{n}$.

	After receiving the trained models and velocities from $n$ training vehicles, the BS firstly calculates the blur level $\mathcal{B}_{r}^{n}$ based on the received velocity of vehicle $n$ in the previous slot, as \cite{sshirmohammadi}\cite{jacortés-Osorio2018}:
	\begin{equation}
		\mathcal{B}_{r}^{n}=\frac{sH}{Q}v_{k,t-1}^{n},
		\label{eq4}
	\end{equation}where $s$ is the focal length, $H$ is the exposure time interval, and $Q$ is pixel units. 
	
	The issue of motion blur increases the difficulty of obtaining high-quality data while protecting privacy. To mitigate the negative impact of blurred images on model training, the proposed DRL-BFSSL algorithm enables to dynamically adjust the weight of blurred images in model aggregation. 
	According to Eq. \eqref{eq4}, we assign smaller weights to the parameters obtained from training vehicles with higher blur levels in the local model, thus reducing the impact of the motion blur and improving the performance of the global model. Then the BS employs a weighted federated algorithm to aggregate the parameters of $N$ models based on the blur level $\mathcal{B}_{r}^{n}$. The expression for the aggregated new global model is
	\begin{equation}
		\theta_{r}^{\text{global}}=\sum_{n=1}^{N}\left[\frac{\left(\sum_{n=1}^{N}\mathcal{B}_{r}^{n}-\mathcal{B}_{r}^{n}\right)C_{k,t}^{n}(p_{k,t}^{n})\theta_{r}^{\text{local},n}}{\sum_{n=1}^{N}\left(\sum_{n=1}^{N}\mathcal{B}_{r}^{n}-\mathcal{B}_{r}^{n}\right)}\right],
		\label{eq133}
	\end{equation}where $\theta_{r}^{\text{global}}$ is the parameters of new global model. 
	
	At the same time, BS calculates the reward $r_{k,t}$ according to Eq. \eqref{eq43}. After getting $r_{k,t}$, state $s_{k,t}$ transitions to $s_{k,t+1}$. BS saves the tuple $(s_{k,t},a_{k,t},r_{k,t},s_{k,t+1})$ to the replay buffer. When storing tuples, if the number of stored tuples is less than the storage capacity, the tuple is directly stored. 
	
	Repeat \textbf{\emph{Step 2}} to \textbf{\emph{Step 4}} until slot $t$ reaches to $S^{\text{max}}$. At this point, if the episode $k$ is divided by $K_u$, the five networks of SAC need to be updated, otherwise, it is not required. After that, episode $k$ moves to episode $k+1$, Slot $t$ reset to $0$. Next, we will introduce the update process of five networks.
	
	In training stage, SAC algorithm optimizes the policy of actor network $\varsigma$ to achieve higher cumulative rewards while also maximizing the policy entropy. Our objective is to find the optimal policy $\pi^\ast(a_{k,t}| s_{k,t})$ of $\pi(a_{k,t}| s_{k,t})$ that maximizes both the long-term discounted rewards and the entropy of the policy simultaneously, i.e., maximum $\mathcal{J}(\pi(a_{k,t}| s_{k,t}))$, which can be expresses as
	\begin{equation}
		\begin{aligned}
			\mathcal{J}(\pi(a_{k,t}| s_{k,t}))=E_{s_{k,t}\sim\mathcal{D}} \sum_{t=1}^{\mathcal{S}^{\text{max}}}\left[ \gamma^{t-1}r_{k,t}\right.\left. + \alpha H[\pi\left(\cdot\middle| s_{k,t}\right)] \right],
			\label{eq44}
		\end{aligned}
	\end{equation}where $\mathcal{D}$ is the distribution of previously sampled states and actions. $\gamma\in[0,1]$ denotes the discount factor. $\pi\left(\cdot\middle| s_{k,t}\right)$ represents the policy after taking all actions in state $s_{k,t}$. $H[\pi\left(\cdot\middle| s_{k,t}\right)]=E[\text{log}\pi\left(\cdot\middle| s_{k,t}\right)]$, and signifies the policy entropy. $\alpha$ measures the weight on policy entropy while maximizing discounted rewards. Moreover, in state $s_{k,t}$, $\alpha$ can dynamically adjust based on the optimal $\alpha^\ast$, i.e.,
	\begin{equation}
		\alpha^\ast = -\arg\min_\alpha E[\alpha\log \pi^\ast(a_{k,t}| s_{k,t}) + \alpha \widetilde{H}],
		\label{eq45}
	\end{equation}where $\widetilde{H}$ represents the dimensions of action $a_{k,t}$.

	The SAC algorithm prefers to achieve efficient and stable decision-making by continuously updating the policy through maximizing cumulative rewards and policy entropy. This is done by adding new data to the experience replay buffer and utilizing target networks. For each iteration of the updating process, randomly sample $\mathcal{M}$ tuples from the replay buffer to form a mini-batch for training. For the $m$-th tuple $(s_m, a_m, r_m, s_{m+1})$ in the mini-batch, where $m \in [1, \mathcal{M}]$, inputting $s_m$ to the actor network to obtain $a_m'$. It is worth noting that this action $a_m'$ is not the same as $a_m$ in the tuple. Thus, the gradient of loss function of $\alpha$ can be calculated as
	\begin{equation}
		\begin{aligned}
			\nabla_{\alpha}\mathcal{J}_{\alpha}
			=\nabla_{\alpha}E\Big[-\frac{1}{\mathcal{M}} \sum_{m=1}^{\mathcal{M}}\left[\alpha\log \pi_{\varsigma}(a_m'|s_m) + \alpha \widetilde{H}\right]^2\Big].
			\label{eq48}
		\end{aligned}
	\end{equation}
	
	Feeding $(s_m,a_m')$ into the two critic networks $\xi_1$ and $\xi_2$, and then produce the corresponding action value functions $Q_{\xi_1}\left(s_m, a_m'\right)$ and $Q_{\xi_2}\left(s_m, a_m'\right)$.  ${Q}_{\xi}\left(s_m, a_m'\right)$ is donated as the smaller between $Q_{\xi_1}\left(s_m, a_m'\right)$ and $Q_{\xi_2}\left(s_m, a_m'\right)$, and can be expressed as
	\begin{equation}
		{Q}_{\xi}\left(s_m, a_m'\right) = \min\left[Q_{\xi_1}\left(s_m, a_m'\right), Q_{\xi_2}\left(s_m, a_m'\right)\right].
		\label{eq47}
	\end{equation} 
	
	Therefore, the the gradient of loss of the actor network parameters can be calculated by
	
	\begin{equation}
		\begin{split}
			&\nabla_\varsigma \mathcal{J}_\varsigma = \nabla_\varsigma \left[ -\frac{1}{\mathcal{M}} \sum_{m=1}^{\mathcal{M}} \left[ \alpha \text{log} \pi_\varsigma \left( a_m^\prime \middle| s_m\right) \right]^2 \right] + \\&\nabla_\varsigma \left[ -\frac{1}{\mathcal{M}} \sum_{m=1}^{\mathcal{M}}\left[{Q}_{\xi}(s_m, a_m^\prime) - \alpha \text{log} \pi_\varsigma \left( a_m^\prime \middle| s_m \right)\right]^2 \right]\\&\times \nabla_\varsigma f_\varsigma(o; s_m),
		\end{split}
		\label{eq46}
	\end{equation}where $o$ is noise drawn from a multivariate normal distribution, such as a spherical Gaussian, and $f_\varsigma(o; s_m)$ is a function used to reparameterize action $a_m^\prime$ \cite{thaarnoja2018}. Finally, the actor network is updated by SGD method.

\begin{algorithm}[t]
	\caption{The process of updating networks of SAC Algorithm}
	\label{al1}
	\KwIn{$\gamma$, $\alpha$, $\varsigma$, $\xi_1$, $\xi_2$, $\varpi_1$, $\varpi_2$, $\mathcal{M}$, $k$, $K_t$}
	
	\SetKwFunction{FUpdate}{UPDATE}
	\SetKwProg{Fn}{Function}{:}{}
	\Fn{\FUpdate{$\gamma$, $\alpha$, $\varsigma$, $\xi_1$, $\xi_2$, $\varpi_1$, $\varpi_2$, $\mathcal{M}$, $k$, $K_t$}}{
		
		Randomly sample $\mathcal{M}$ tuples from replay buffer
		
		Update $\alpha$ according to Eq. (\ref{eq48});
		
		Update $\varsigma$ according to Eq. (\ref{eq46});
		
		Update $\xi_1$ and $\xi_2$ according to Eq. (\ref{eq65}) and Eq. (\ref{eq49});
		
		\If{$k \mod K_t = 0$}{
			Update $\varpi_1$ and $\varpi_2$ according to Eq. (\ref{eq51}) and Eq. (\ref{eq52});

		}
	}
\end{algorithm}
	
\begin{algorithm}[t]
	\caption{The DRL-BFSSL Algorithm}
	\label{al2}
	\KwIn{$\theta^0$, $\varsigma$, $\xi_1$, $\xi_2$, $\varpi_1$, $\varpi_2$}
	Initialize round $r \gets 0$, episode $k \gets 0$\;
	\For{episode $k$ from $1$ to $K^{\text{max}}$}{
		slot $t \gets 0$\;
		\For{slot $t$ from $1$ to $S^{\text{max}}$}{
			$r \gets (k-1) \cdot S^{\text{max}} + t$\;
			Obtain state $s_{k,t}$ based on distance and velocity\;
			Get action $a_{k,t}$ through SAC algorithm\;
			\For{$n = 1, 2, \dots, N$ in parallel}{
				Download global model $\theta_{r-1}^{\text{global}}$, $p_{k,t}^{n}$, and $f_{k,t}^{n}$\;
				$\theta_{r}^{{\text{local}},n} \gets \text{LOCAL}(\theta_{r-1}^{\text{global}}, p_{k,t}^{n}, f_{k,t}^{n})$\;
			}
			\If{$r = 1$}{
				Vehicle $n \in [1, N]$ uploads the local model $\theta_{r}^{{\text{local}},n}$ and $v_{k,t}^{n}$ to BS with download transmission power $p_{k,t}^{n}$, along with velocity $v_{0}^{n}$\;
			}
			\Else{
				Vehicle $n \in [1, N]$ uploads the local model $\theta_{r}^{{\text{local}},n}$ and $v_{k,t}^{n}$ to BS with download transmission power $p_{k,t}^{n}$\;
			}
			BS aggregates received local models based on $\mathcal{B}_{r}^n$ and updates $\theta_{r}^{\text{global}}$ according to Eq. \eqref{eq133}\;
			Save $\theta_{r}^{\text{global}}$ to global model\;
		}
		\If{episode $k$ mod $K_u$ = 0}{
			\text{UPDATE}($\gamma$, $\alpha$, $\varsigma$, $\xi_1$, $\xi_2$, $\varpi_1$, $\varpi_2$)\;
		}
	}
	\SetKwFunction{FLocal}{Local}
	\SetKwProg{Fn}{Function}{:}{}
	\Fn{\FLocal{$\theta_{r-1}^{\text{global}}, p_{k,t}^{n}, f_{k,t}^{n}$}}{
		\KwIn{$\theta_{r-1}^{\text{global}}, p_{k,t}^{n}, f_{k,t}^{n}$}
		Set the parameter of global model $\theta_{r-1}^{\text{global}}$ to local model $\theta_{r}^{{\text{local}},n}$\;
		Calculate iteration value $\mathcal{N}_{k,t}^{n}$ according to Eq. (\ref{eq20}) and Eq. (\ref{eq111})\;
		\For{each iteration $\psi_{k,t}^n$ from $1$ to $\mathcal{N}_{k,t}^{n}$}{
			Prepare local model $\theta_{r}^{{\text{local}},n}$\;
			\For{image $i$ from $1$ to $\mathcal{Z}$}{
				Calculate $q_{r}^{i,n}$, $k_{r}^{i,n}$, and $k_{r}^{j,n}$ according to Eq. (\ref{eq5}) - Eq. (\ref{eq77})\;
			}
			Calculate the average loss of $\mathcal{Z}$ images according to Eq. (\ref{eq7}) - Eq. (\ref{eq9}), and update $\theta_{r}^{{\text{local}},n}$ according to SGD algorithm in Eq. (\ref{eq12})\;
		}
		\KwOut{$\theta_{r}^{{\text{local}},n}$}
	}
\end{algorithm}
	For updating critic networks $\xi_1$ and $\xi_2$, firstly, we use $(s_m,a_m)$ as input and feed it separately into the critic networks $\xi_1$ and $\xi_2$, and obtain the action values pairs $Q_{\xi_1}\left(s_m,a_m\right)$ and $Q_{\xi_2}\left(s_m,a_m\right)$, respectively. Additionally, input $s_{m+1}$ into the actor network to obtain $a_{m+1}$, and represented as $\text{log}\pi_\varsigma^{m+1}(a_{m+1}|s_{m+1})$ which is to convert the policy parameter space into real number space, and this transformation helps with the computational efficiency and numerical stability of the optimization algorithm. Finally, use $\left(s_{m+1},a_{m+1}\right)$ as input for the two target networks $\varpi_1$ and $\varpi_2$, and output the action value pairs $Q_{\varpi_1}\left(s_{m+1},a_{m+1}\right)$ and $Q_{\varpi_2}\left(s_{m+1},a_{m+1}\right)$. We take the minimum of the two, denoted as $Q_\varpi\left(s_{m+1},a_{m+1}\right)$. From this, we can calculate the target value $\hat{Q}(s_{m+1},a_{m+1})$ as
	\begin{equation}
		\begin{split}
			\hat{Q}(s_{m+1}, a_{m+1}) =& Q_\varpi(s_{m+1}, a_{m+1}) \\
			&- \alpha \log \pi_\varsigma^{m+1}\left(a_{m+1} \middle| s_{m+1}\right),
		\end{split}
		\label{eq65}
	\end{equation}and the gradient of loss of critic networks $\xi_1$ and $\xi_2$ can be calculated as, respectively
	\begin{equation}
		\begin{split}
			\nabla_{\xi_1}\mathcal{J}_{\xi_1} = \nabla_{\xi_1}\Big[ -\frac{1}{\mathcal{M}} \sum_{m=1}^{\mathcal{M}}{[ r_m + \gamma \hat{Q}(s_{m+1}, a_{m+1})} \\
			\quad - Q_{\xi_1}(s_m, a_m) ]^2\Big],  \xi \in{\xi_1,\xi_2}
		\end{split}
		\label{eq49}
	\end{equation}

	Finally, SGD method is employed to update the two critic networks. We also need to update both target networks for every $K_t$ iterations during this process. The updates for the two target networks are as follows
	\begin{equation}
		\varpi_1=\delta_1\ \xi_1+(1-\delta_1)\varpi_1,
		\label{eq51}
	\end{equation}
	\begin{equation}
		\varpi_2=\delta_2\ \xi_2+(1-\delta_2)\varpi_2,
		\label{eq52}
	\end{equation} where $\delta_1$ and $\delta_2$ are constants much smaller than $1$. The pseudocode of the process of updating SAC networks is shown in Algorithm \ref{al1}. 
	
	When episode $k$ reaches to $K^{\text{max}}$ and slot $t$ reached to $S^{\text{max}}$, we get a global model and an optimal parameter of actor network $\varsigma$, denoted as $\varsigma^{\ast}$. The pseudocode of complete algorithm is shown in Algorithm \ref{al2}.

\begin{figure*}[t]
	\centering
	\subfloat[CIFAR-10 IID]{%
		\includegraphics[width=0.32\textwidth]{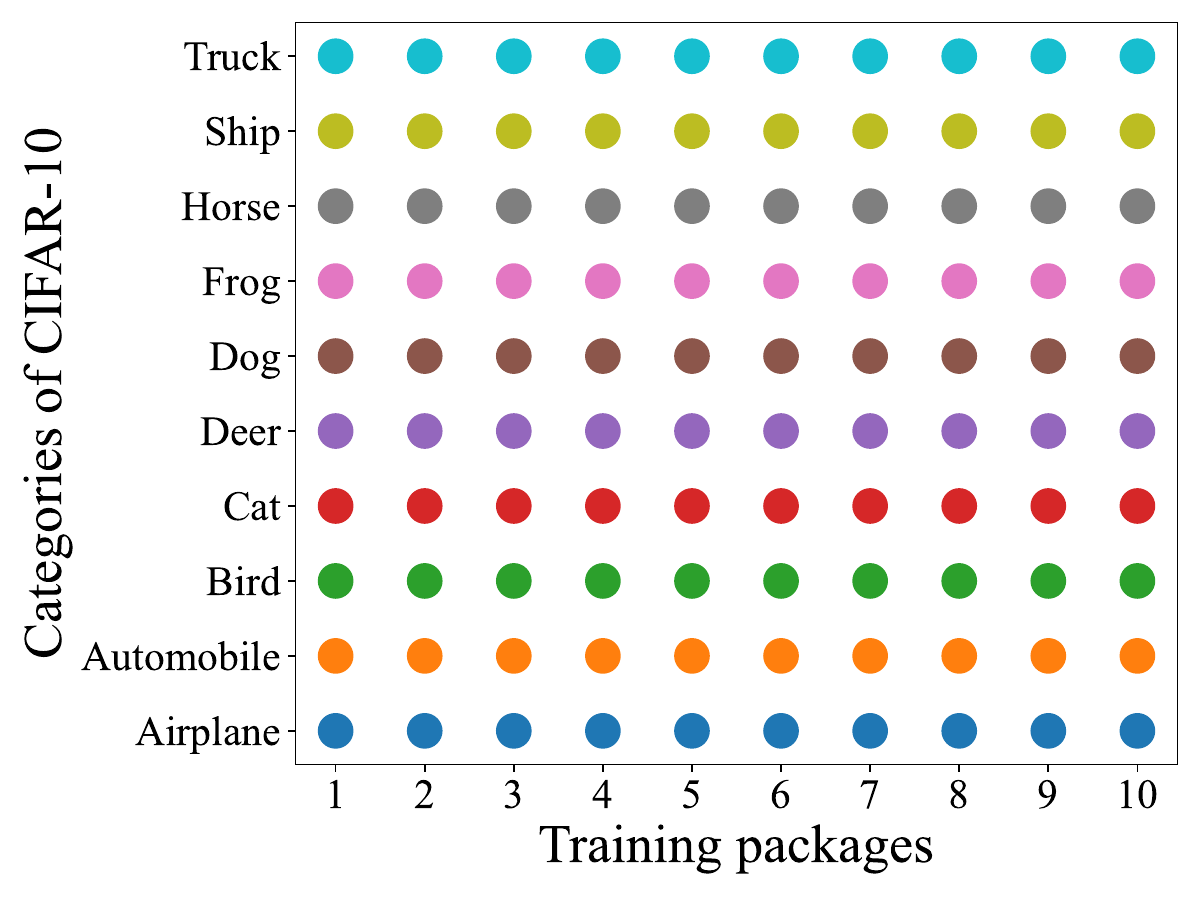}
		\label{fig6a}
	}\hfill
	\subfloat[CIFAR-10 Non-IID, $\mu=0.1$]{%
		\includegraphics[width=0.32\textwidth]{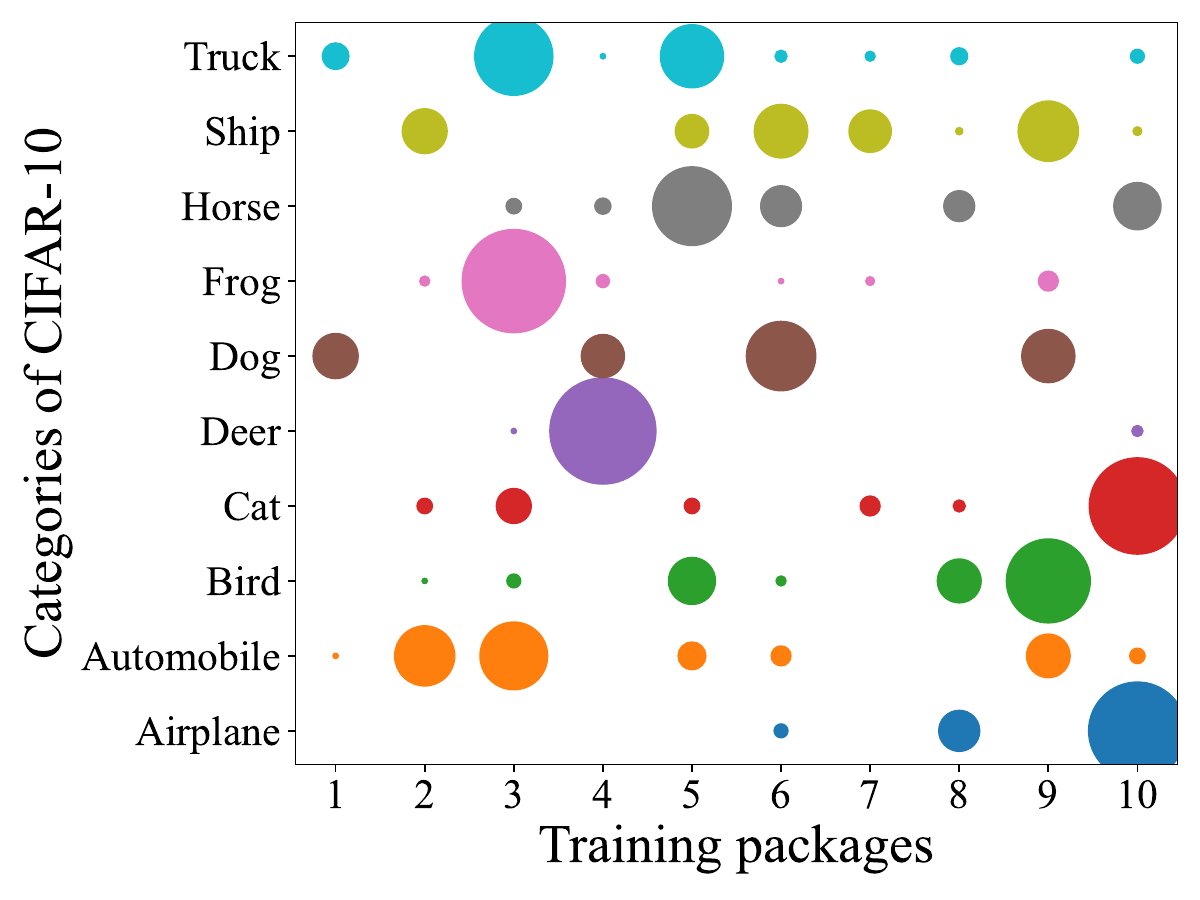}
		\label{fig6b}
	}\hfill
	\subfloat[CIFAR-10 Non-IID, $\mu=1$]{%
		\includegraphics[width=0.32\textwidth]{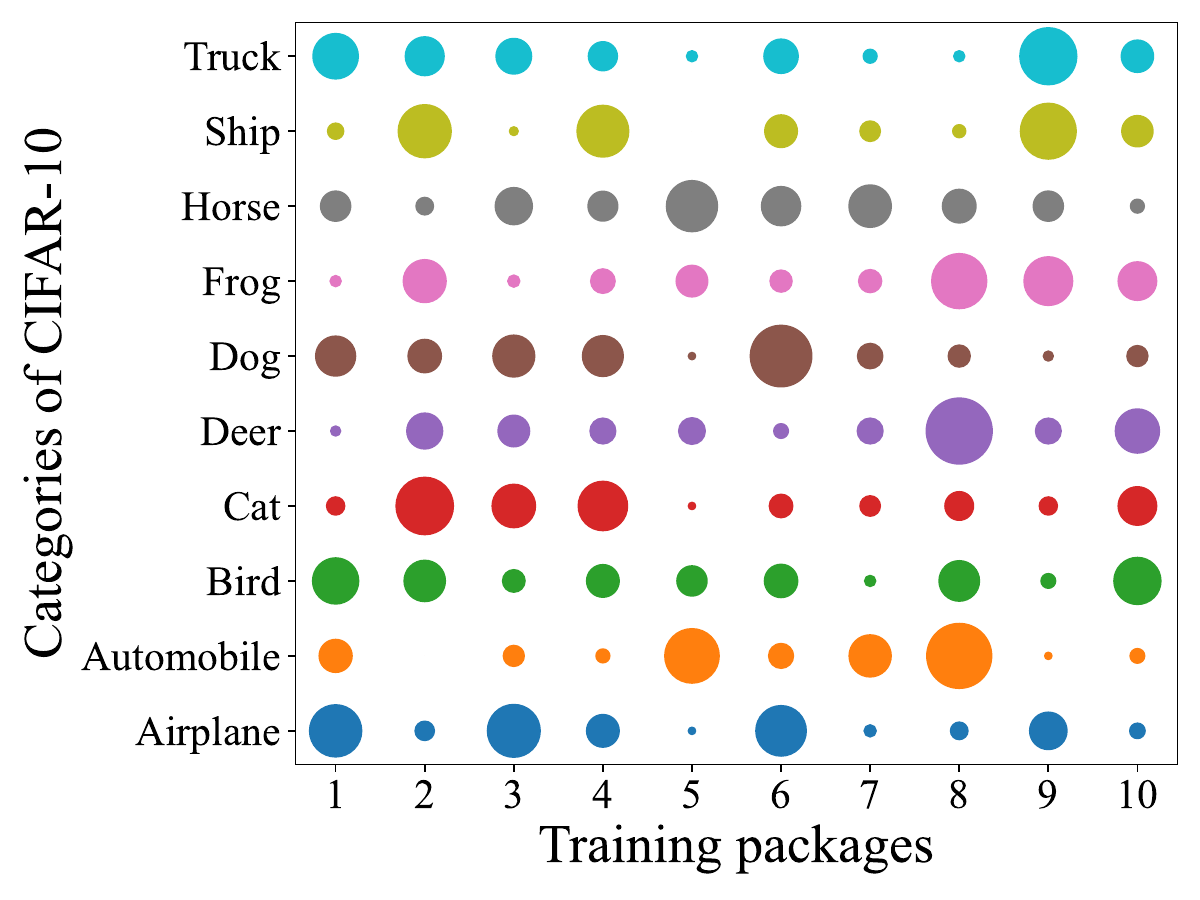}
		\label{fig6c}
	}
	\caption{Data category distribution plot}
	\label{fig6}
\end{figure*}

	\section{Simulation}
	\label{sec6}
	
	Python 3.10 is utilized to conduct the simulations, which are based on the scenarios outlined in the Section \ref{sec3}. The simulation parameters are detailed in Table \ref{tab1}.
	
	\begin{table}[htbp]
		\centering
		\caption{Hyper-parameter}
		\label{tab1}
		\begin{tabular}{|c|c|c|c|}
			\hline
			\textbf{Parameter} & \textbf{Value} & \textbf{Parameter} & \textbf{Value} \\
			\hline						  
			$m$ & 0.023 & $\epsilon_\tau$ & 0.2 \\
			$\tau_\alpha$ & 0.1 & $\tau_\beta$ & 1 \\
			$\mathcal{K}$ & 512 & $\gamma$ & 0.99 \\
			$\delta_1/\delta_2$ & 0.001 & $\text{Initial learning rate}$ & 0.06 \\
			$K^{\text{max}}$ & 1000 & $S^{\text{max}}$ & 100 \\
			$Z$ & 11.2M & $B^U$ & $2\times{10}^6Hz$ \\
			$N_0$ & -114dB & $\xi$ & 8 \\
			$\kappa$ & ${10}^{-27}$ & $\text{momentum of SGD}$ & 0.9 \\
			$f^{\text{min}}$ & $5\times{10}^7Hz$ & $f^{\text{max}}$ & $4\times{10}^8Hz$ \\
			$p^{\text{min}}$ & 5dB & $p^{\text{max}}$ & 200dB \\
			$T$ & 0.5s & $t_{\text{max}}^{\text{Trans}}$ & 0.02s \\
			$\lambda_1$ & 0.7 & $\lambda_2$ & 0.3 \\
			$v^{\text{min}}$ & 60km/h & $v^{\text{max}}$ & 150km/h\\
			$\mu$ & 0.5 & $\sigma^2$ & 8\\
			$\vartheta_1$ & 0.05 & $\vartheta_2$ & $5\times{10}^{-5}$ \\
			$\mathcal{M}$ & 256 & $p_j$ & 128 \\
			$L$ &6& $p_i$ & 512 \\
			$K_u$ & 2 & $K_t$ & 80 \\	
			$\varepsilon_1/\varepsilon_2/\varepsilon_3$ & 0.3/0.3/0.4 &&\\
					
			\hline
		\end{tabular}
	\end{table}

	We employ a modified ResNet-18, with a fixed dimension of $p_j$, as the backbone model and utilize SGD with momentum as the optimizer \cite{3}. Momentum SGD is characterized by its strong randomness, rapid convergence, and ease of implementation, making it well-suited for optimizing the backbone model. Additionally, inspired by the concept of Cosine Annealing, we gradually reduce the learning rate at different stages of training to enhance the model's training effectiveness.

	\subsection{Datasets}	
	In intersection-related scenes, the categories of objects are relatively limited, and images of the same category are frequently collected. Therefore, CIFAR-10 is selected. In CIFAR-10, the training datasets comprises 50,000 images distributed across 10 distinct categories, each consisting of 5,000 images. The testing datasets comprises 10,000 images. We also establish two types of data distribution: IID, and Non-IID.

	\begin{figure}
		\center
		\includegraphics[scale=0.38]{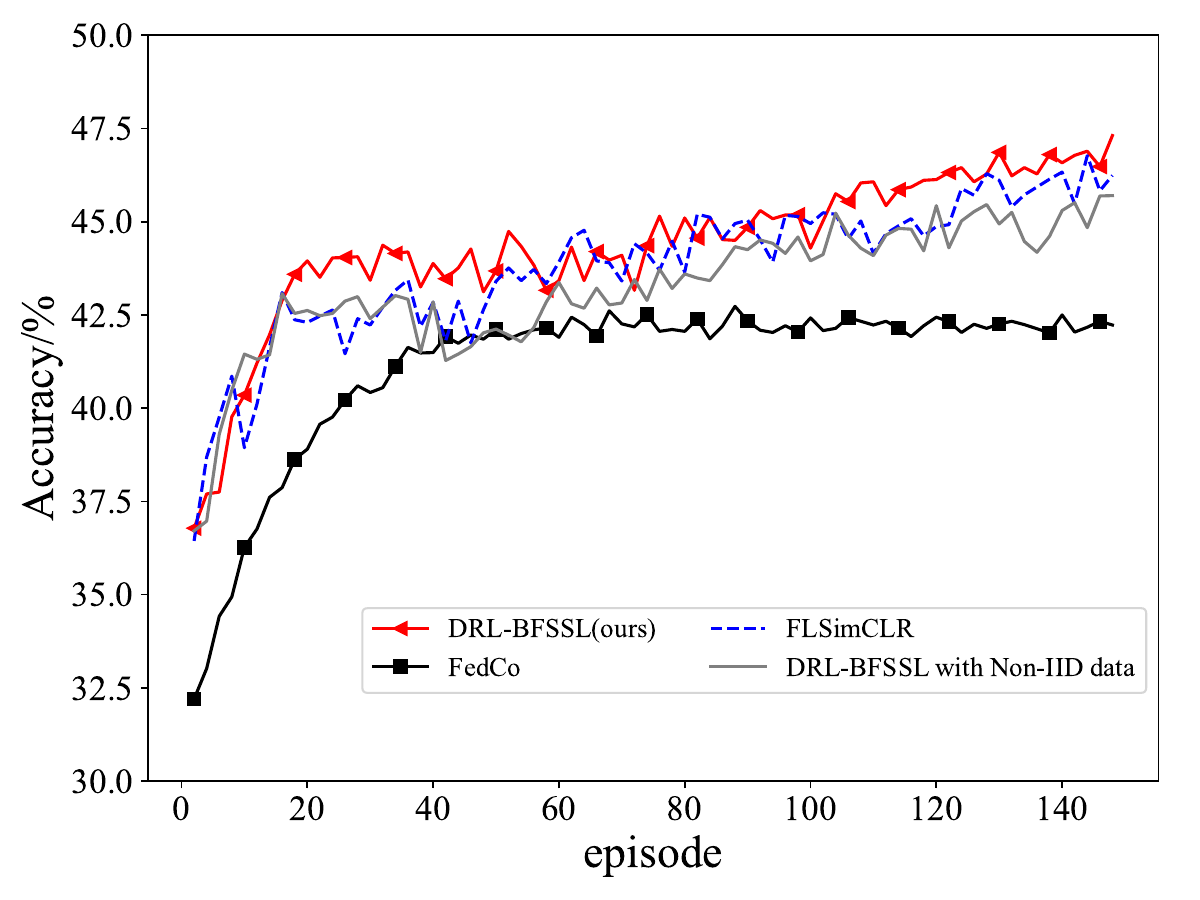}
		\caption{Comparison of DRL-BFSSL with Other Methods}
		\label{fig7}
	\end{figure}
\begin{figure*}[!t]
	\centering
	\subfloat[The accuracy of Top1]{%
		\includegraphics[width=0.32\linewidth]{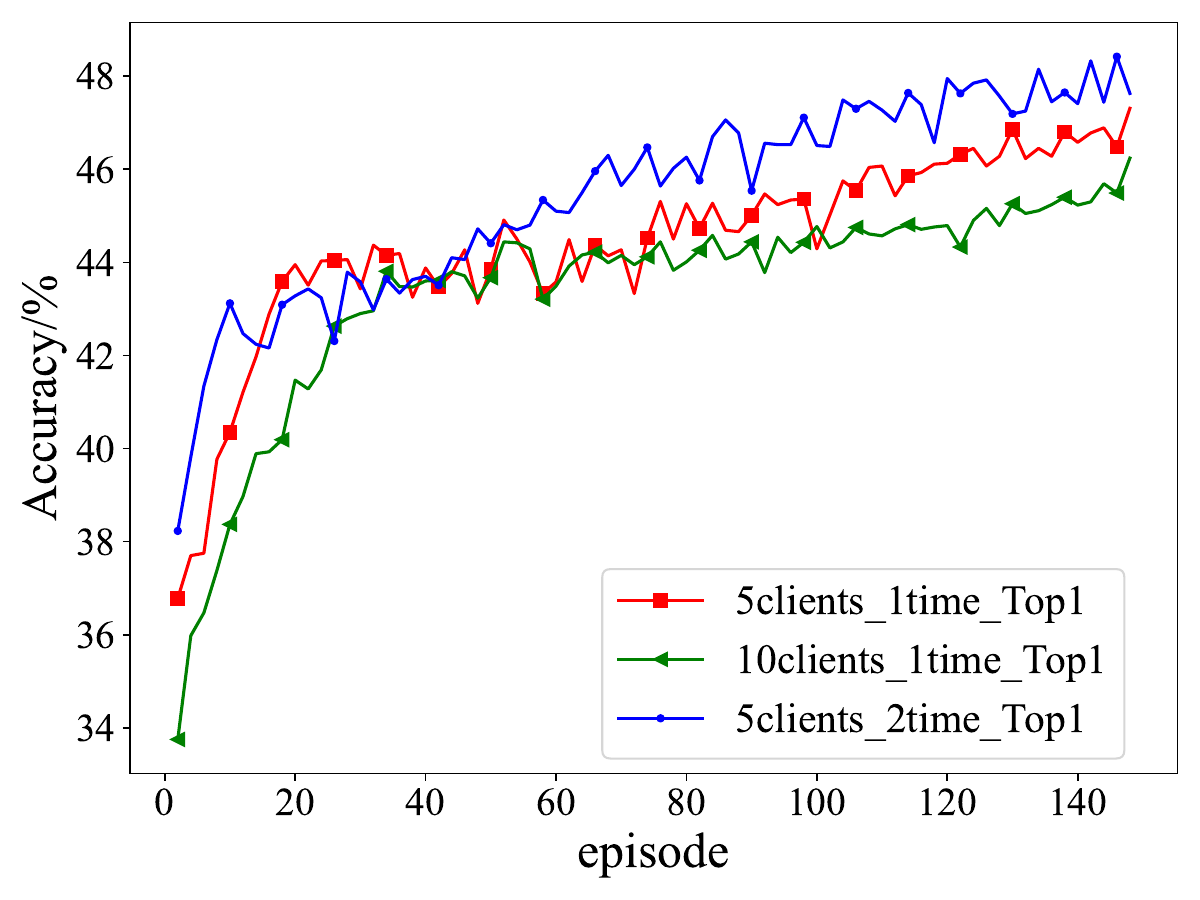}
		\label{fig8a}
	}\hfill
	\subfloat[The accuracy of Top5]{%
		\includegraphics[width=0.32\linewidth]{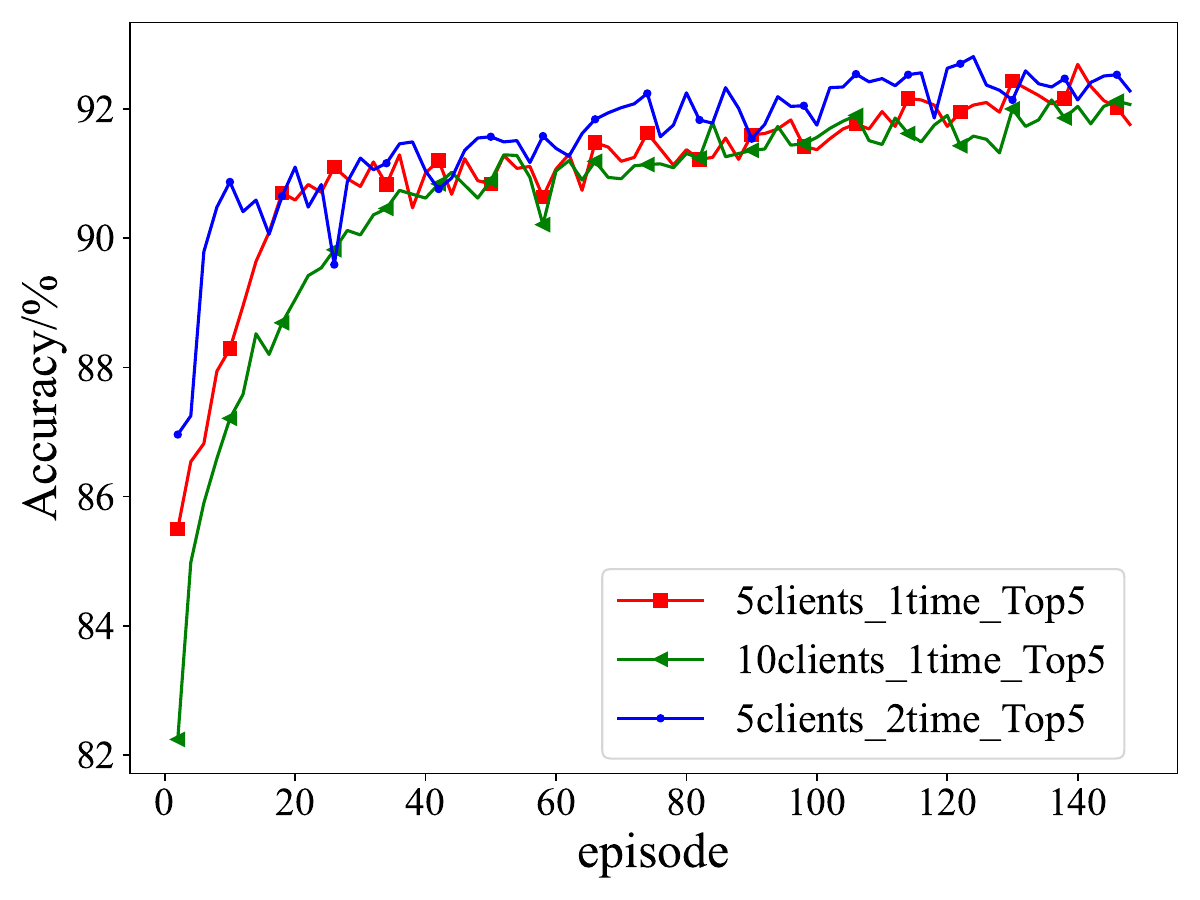}
		\label{fig8b}
	}\hfill
	\subfloat[Loss function]{%
		\includegraphics[width=0.32\linewidth]{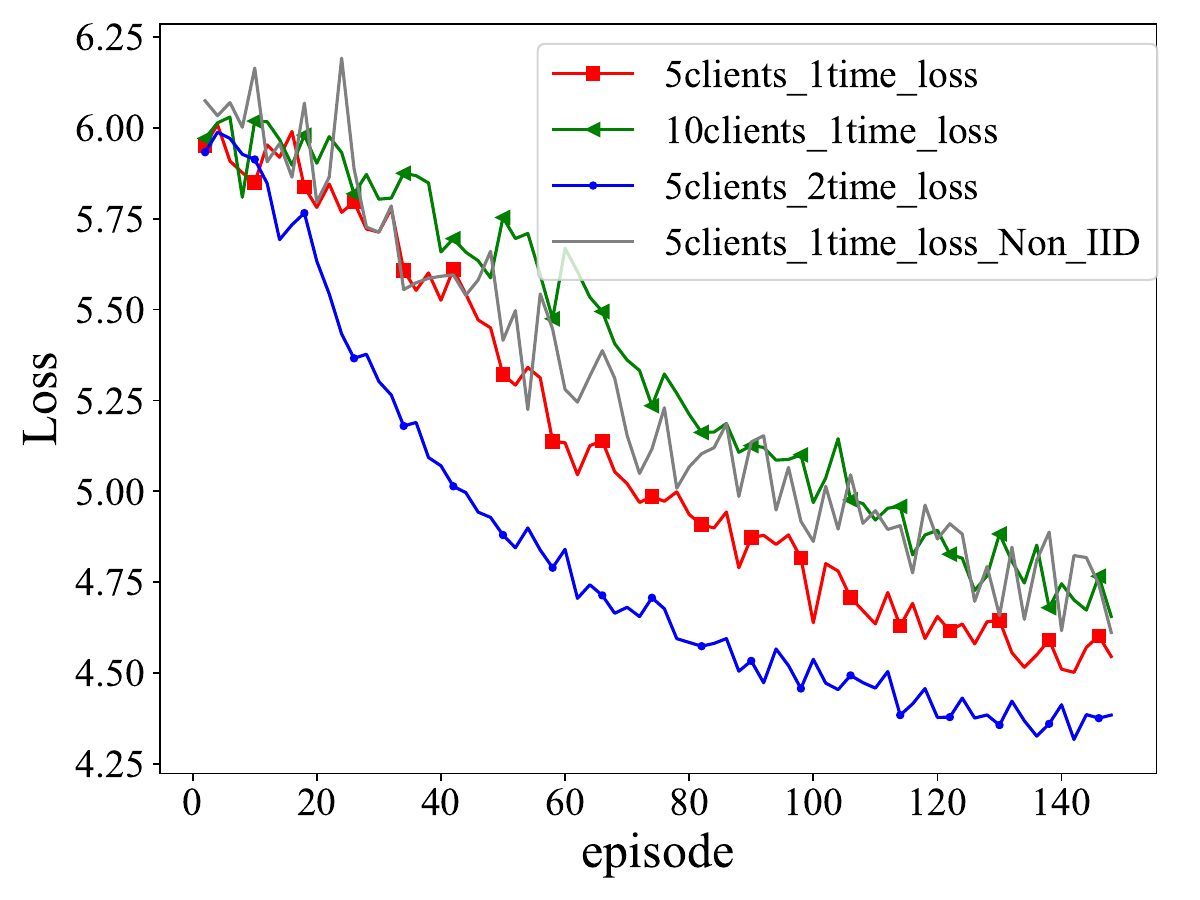}
		\label{fig8c}
	}
	\caption{Comparison of accuracy and loss}
	\label{fig8}
\end{figure*}

	\subsubsection{ \textbf{\emph{IID}}}
	Data obeying an IID are mutually independent and share the same distribution. Simultaneously, the statistical properties among data are similar, facilitating the generalization of the model. We uniformly allocate the 50,000 image data from CIFAR-10 to 95 training packages, ensuring that each training package has at least 520 images available. In Fig. \ref{fig6}, we show the distribution of image data categories on 10 training packages. As shown in Fig. \ref{fig6a}, the quantity of each image data category on each vehicle is equal, indicating a uniform distribution state.

	\subsubsection{\textbf{\emph{Non-IID}}}
	 IID offers theoretical convenience, but in practical scenarios, data satisfied with IID is rare. Therefore, it is beneficial to evaluate the model's training performance under Non-IID conditions for transferring the model to real-world scenarios \cite{khsieh2020}. As depicted in Fig. \ref{fig6b} and Fig. \ref{fig6c}, scenarios with Dirichlet distribution parameters $\alpha$ set to 0.1 and 1 are respectively shown. It can be observed that as $\alpha$ decreases, the gap between different categories increases, while a larger $\alpha$ makes the distribution closer to IID. We set the Dirichlet distribution parameter $\alpha$ to 0.1 to simulate Non-IID data in IoV, aiming to simulate situations where each training vehicle, due to limited perspectives and environmental constraints, collects images with uneven category distributions. 
	
	\subsection{Testing settings}	
 We rank the predicted labels based on their probabilities from highest to lowest. If the most probable predicted label (i.e., the top label) matches the true label, the prediction is considered correct, and this is referred to as the Top1 accuracy. If the true label is among the top five predicted labels, the prediction is also considered correct, and this is referred to as the Top5 accuracy. 
	For testing the performance of SAC algorithm, we employ the output $\varsigma^{\ast}$ to obtain $\pi^\ast(a_{k,t}| s_{k,t})$.
	Each simulation is conducted for three times, and the final result is the average of these three simulations.

	We first test the classification accuracy of the global model obtained under different conditions.  We assume that each episode consists of 95 training packages, with each vehicle selecting one training package for each training session. When all training packages are used up, one episode is finished.
	
			\begin{table}[htbp]
		\centering
		\caption{Parameters for PSO algorithm}
		\label{tab2}
		\begin{tabular}{|c|c|}
			\hline
			\textbf{Parameter} & \textbf{Value} \\
			\hline						  
			Maximum iterations & 100  \\
			Inertia weight & 0.2  \\
			Personal learning coefficient & 0.1 \\
			Social learning coefficient & 0.1 \\
			Positions bounds & $[0.0001,1]$ \\			
			\hline
		\end{tabular}
	\end{table}	
	
	\subsection{Comparative algorithms Analysis}
	\subsubsection{Particle Swarm Optimization (PSO)}
	A population-based optimization algorithm that simulates the behavior of bird flocks or fish schools in search of food. Each particle moves in the solution space using position and velocity, where position represents a solution and velocity determines the direction of movement \cite{PSO}. And the detailed parameter settings are shown in Table \ref{tab2}.
	\subsubsection{Heuristic Gibbs Greedy Algorithm (HGGA)}
	An optimization algorithm that combines heuristic methods and greedy strategies, primarily used for combinatorial optimization problems. It generates candidate solutions through heuristic rules and Gibbs sampling, prioritizing decisions that significantly impact solution quality to accelerate the search process \cite{HGGA}. 
	\subsubsection{Hierarchical Genetic Recombination Algorithm (HGRA)}
	An optimization algorithm that combines heuristic methods and random search, primarily used for solving complex combinatorial optimization problems. It introduces randomness and heuristic rules during the search process to avoid getting stuck in local optima and improve global search capability \cite{HGRA}.
	\subsubsection{DDPG}
	An DRL algorithm suitable for continuous action spaces, combining the concepts of DQN and the Actor-Critic architecture. It employs an Actor-Critic structure where the Actor selects actions and the Critic evaluates their values. DDPG updates the policy deterministically and uses experience replay and target networks to stabilize the training process.
	\subsubsection{Twin Delayed Deep Deterministic Policy Gradient (TD3)}
	An improved version of DDPG designed to address the overestimation bias present in DDPG. It enhances stability and performance by introducing twin Critic networks, target policy smoothing, and delayed updates. TD3 reduces noise interference during policy updates, demonstrating better convergence and robustness, particularly for complex continuous control tasks.
	\subsubsection{Proximal Policy Optimization (PPO)}
	A policy optimization algorithm that is a simplified version of Trust Region Policy Optimization (TRPO), balancing performance and stability. It stabilizes the training process by clipping policy updates, limiting the change between old and new policies. PPO is applicable to both discrete and continuous action spaces, and its simplicity and superior performance have led to its widespread use in various reinforcement learning tasks.

	\subsection{Complexity analysis}
	Regarding the complexity analysis of the DRL-BFSSL algorithm, which combines SAC and SSL, the computational complexity mainly arises during model training.
	\subsubsection{Complexity analysis of SAC}
	SAC includes both a policy network (actor) and two value networks (critic). Each network has $L$ layers (4 hidden layers and 2 output layers), and the number of parameters in each layer is $p_i$. The computational complexity for a single forward and backward pass is $\mathcal{O}(\sum_{i=1}^{L}p_i)$. Let $\mathcal{M}$ be the batch size and $K$ be the number of multi-step updates. The computational complexity considering the batch size and multi-step updates is $\mathcal{O}(\mathcal{M}K\sum_{i=1}^{L}p_i)$.
	
	\subsubsection{Complexity analysis of SSL}
	$\mathcal{M}$ be the batch size, and the number of parameters in each layer is $p_j$. The computational complexity of SSL is $\mathcal{O}(\mathcal{M}^2p_j)$.
	
	\subsubsection{Complexity analysis of DRL-BFSSL}
	Since DRL-BFSSL algorithm separates SSL training and resource allocation into two parallel-running components, the overall computational complexity of the proposed algorithm can be expressed as $\mathcal{O}[\text{max}(\mathcal{M}K\sum_{i=1}^{L}p_i,\mathcal{M}^2p_j)]$.

	\subsection{Simulation results}
	In Fig. \ref{fig7}, we compared our proposed DRL-BFSSL algorithm with FedCo \cite{sweiFedCo} and FLSimCLR \cite{njahan2023}, under the same local iterations. We also conducted simulations on Non-IID datasets based on our proposed DRL-BFSSL algorithm. It can be seen that our proposed DRL-BFSSL algorithm achieves the highest classification accuracy, followed by FLSimCLR, with FedCo having the lowest classification accuracy. At the same time, although the classification accuracy of DRL-BFSSL on Non-IID data is lower than on IID datasets, it still remains higher than that of FedCo. 
	
	The lower classification accuracy of FedCo can be attributed to its method described in FedCo, where in each round $r$, each training vehicle uploads all stored $k$ values to BS to form a new global queue for updating the global dictionary. However, updating $k$ values from different training vehicles disrupts the Negative-Negative consistency requirement in MoCo, leading to reduced accuracy for these methods. The classification accuracy of the FLSimCLR algorithm is lower than that of our proposed algorithm, primarily because FLSimCLR, which is based on SimCLR, relies heavily on a large number of negative samples during training. However, in our simulation, given the limited number of images, which did not provide an adequate number of negative samples. Consequently, the classification accuracy of FLSimCLR was lower.	Additionally, the slightly lower performance of DRL-BFSSL on Non-IID data compared to IID datasets is primarily due to the inherent challenges in handling heterogeneous data, which affect overall consistency and learning efficiency in federated settings. From another perspective, the stable performance of DRL-BFSSL on Non-IID datasets also indicates that our proposed algorithm has good adaptability and robustness in the face of data heterogeneity that are more common in practical applications.
	
	In Fig. \ref{fig8}, we compared the Top1 and Top5 classification accuracy and the corresponding trends in the loss function under varying numbers of vehicles and local iterations per round. Comparing the red and green lines in Fig. \ref{fig8a}, it shows that model accuracy improves as fewer vehicles participate. This indicates that, since the total number of training vehicles is fixed for each training episode, a decrease in the number of vehicles participating in each round of DRL-BFSSL leads to an increase in the number of communication rounds with BS. This, in turn, results in more communication rounds between the training vehicles and BS, and more communication rounds between vehicles and BS lead to better model performance. The red and blue lines represent scenarios with same training vehicles participating in DRL-BFSSL per round. The red line indicates that training vehicles perform one local iteration, while the blue line indicates two local iterations. It can be observed that the results of two local iterations surpass those of one iteration. This indicates that increasing the number of local iterations leads to higher classification accuracy. We can also observe that when 10 vehicles participate in each round of DRL-BFSSL, the initial accuracy is the lowest, and as the training rounds increase, it gradually becomes comparable to the accuracy achieved with 5 vehicles. In the early stages of training, the local models may not be fully trained, their performances are not stable. At this point, aggregating more models could combine the errors of these inadequately trained models into the global model, leading to poor performance. However, as the number of episodes increases and more data is used, the quality of model training stabilizes, and the impact of data heterogeneity on model performance becomes less significant. 
	
	In Fig. \ref{fig8b}, the Top5 accuracy under the same training conditions as Fig. \ref{fig8a} is displayed. The trend of the curves is similar to that in Fig. \ref{fig8a}, but the gaps between the curves are narrowing, and the accuracy is significantly improved. This suggests that the model becomes more robust and comprehensive when considering more candidate classes. This can be interpreted as the model demonstrating greater tolerance in multi-class classification, being more likely to include the correct classes and better handling the classification uncertainty of certain samples.
	\begin{figure}
		\center
		\includegraphics[scale=0.38]{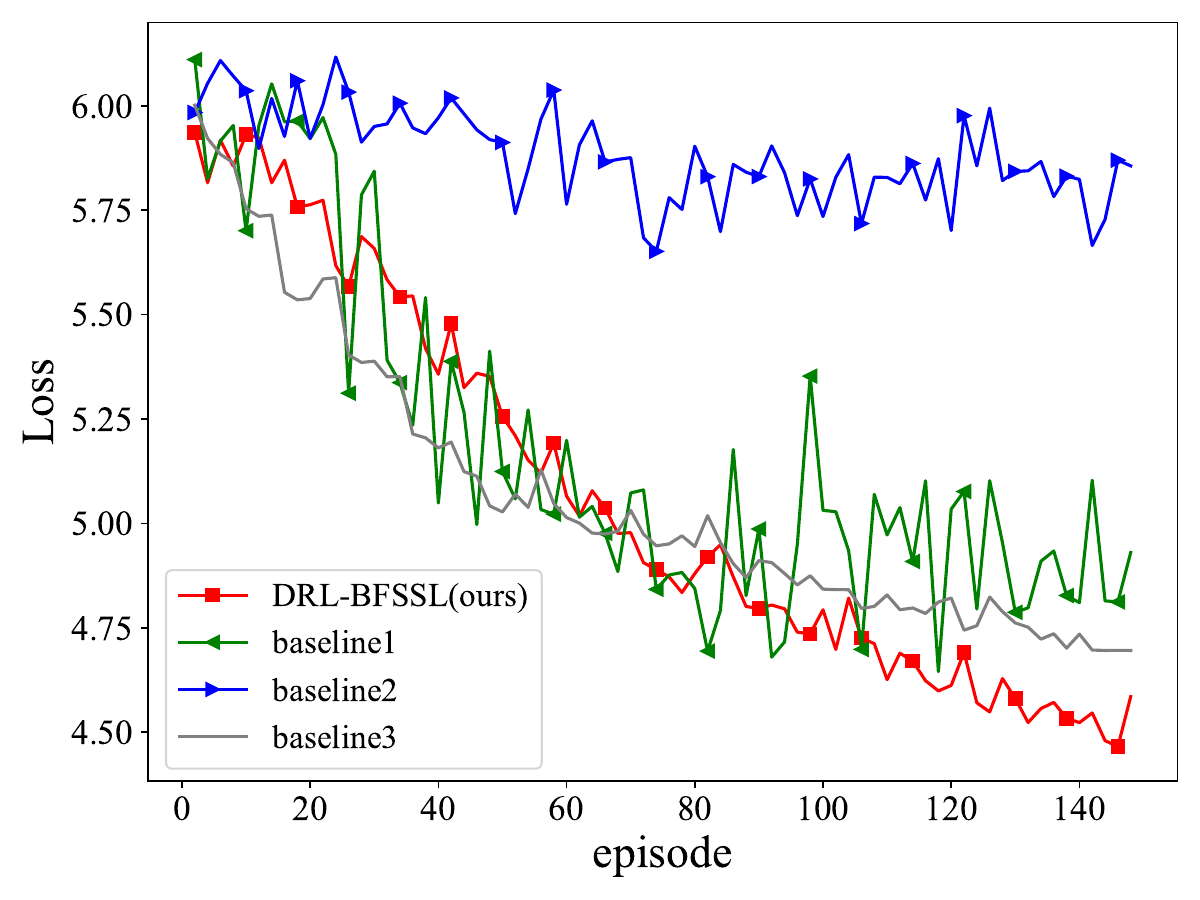}
		\caption{The loss of models aggregated with different weights}
		\label{fig10}
	\end{figure}

		\begin{figure}
		\center
		\includegraphics[scale=0.41]{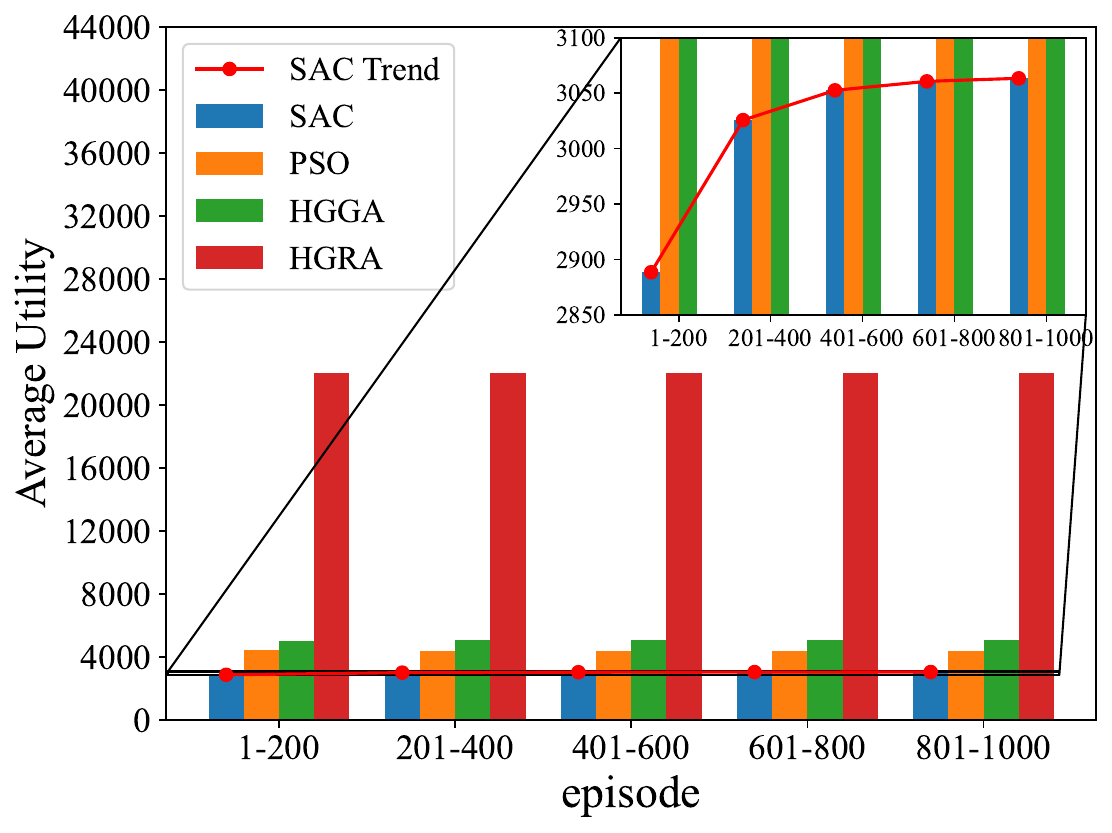}
		\caption{Average utility between SAC and other improved traditional algorithms}
		\label{fig100}
	\end{figure}
	
	As shown in Fig. \ref{fig8c}, we compared the loss curves of the trained models in Fig. \ref{fig8a} on IID and Non-IID datasets. It can be observed that the loss function of each simulation exhibits a decreasing trend. Moreover, when 5 vehicles participate in each round of training with each vehicle conducting two local iterations, the loss decreases the fastest. That's because multiple local iterations help the model better learn data features, allowing it to achieve lower loss values in a shorter time. At the same time, compared to the simulations on IID datasets, the loss function of the model trained on Non-IID datasets exhibits significant fluctuations in the initial stage. However, as the number of episodes increases, the trend of the loss function gradually becomes similar to that on IID datasets.
	This phenomenon can be partially attributed to the characteristics of Non-IID datasets. Due to the uneven distribution and heterogeneity of data on Non-IID datasets, the model faces greater challenges in initial stage of training. As training progresses, the model gradually adapts to the data heterogeneity, leading to a reduction in the fluctuation of the loss function. From the perspective of curve variability, the loss function curve exhibits certain fluctuations. This phenomenon is due to models being trained across multiple decentralized vehicles, which can introduce variability due to differences in local data heterogeneity, training conditions, and the frequency of model updates.
	
\begin{figure*}[!t]
	\centering
	\subfloat[Optimization problem Comparison]{%
		\includegraphics[width=0.32\linewidth]{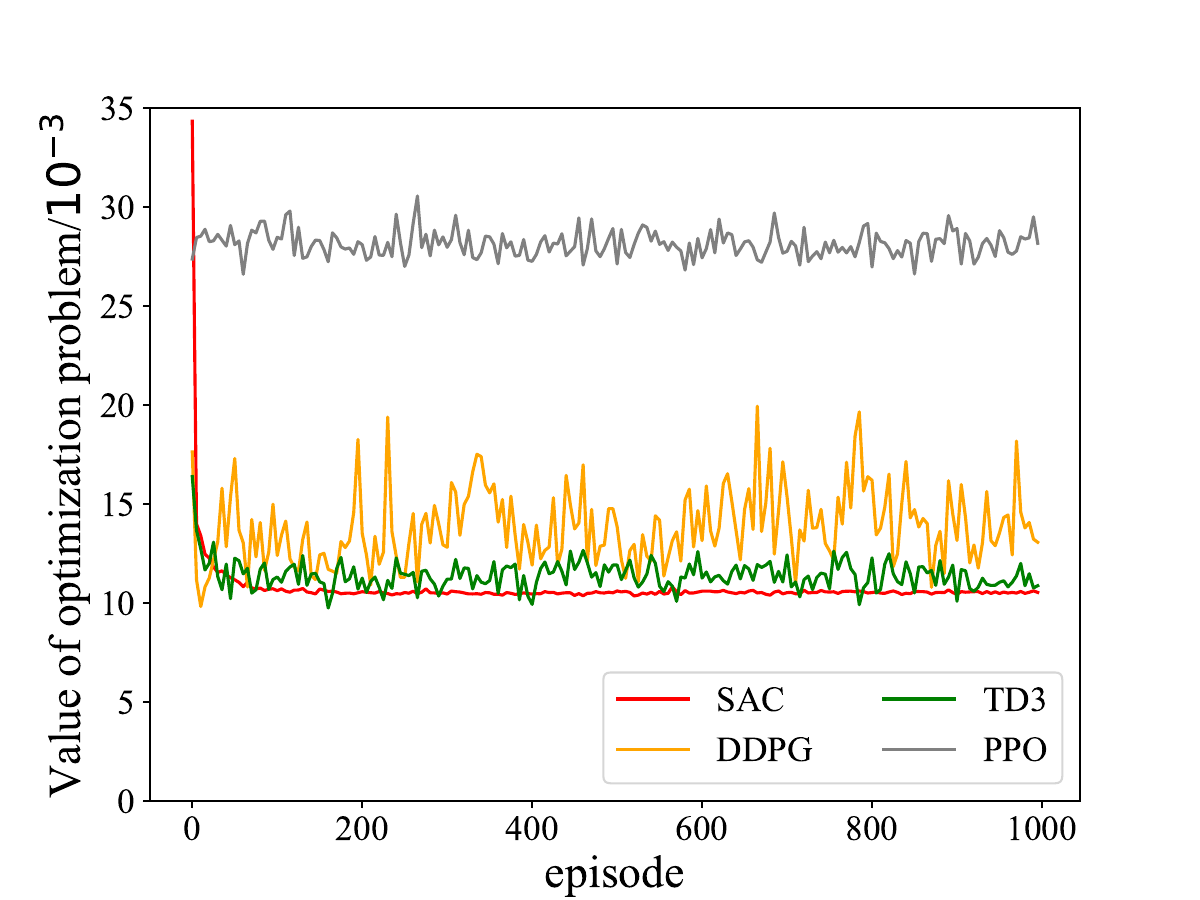}
		\label{fig14a}
	}\hfill
	\subfloat[Calculation Comparison]{%
		\includegraphics[width=0.32\linewidth]{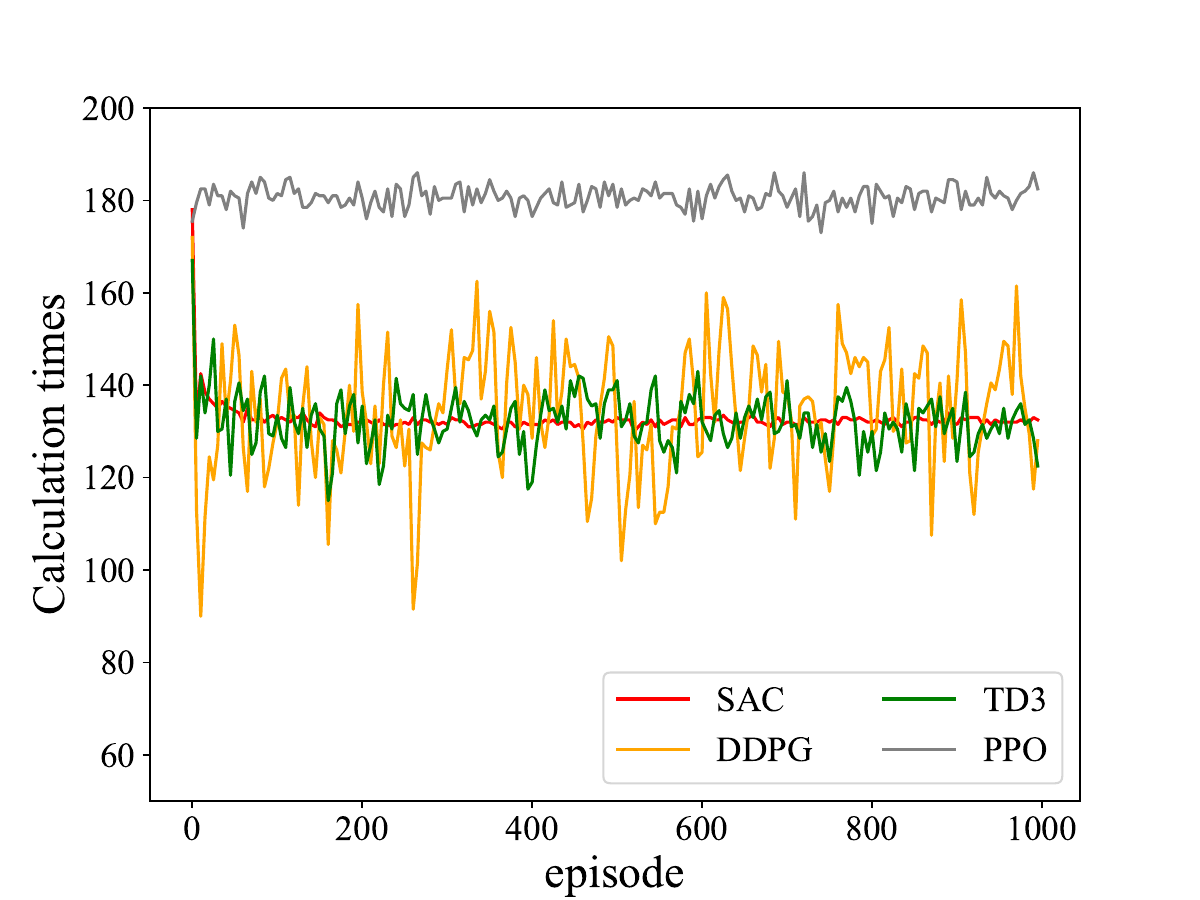}
		\label{fig14b}
	}\hfill
	\subfloat[Reward Comparison]{%
		\includegraphics[width=0.32\linewidth]{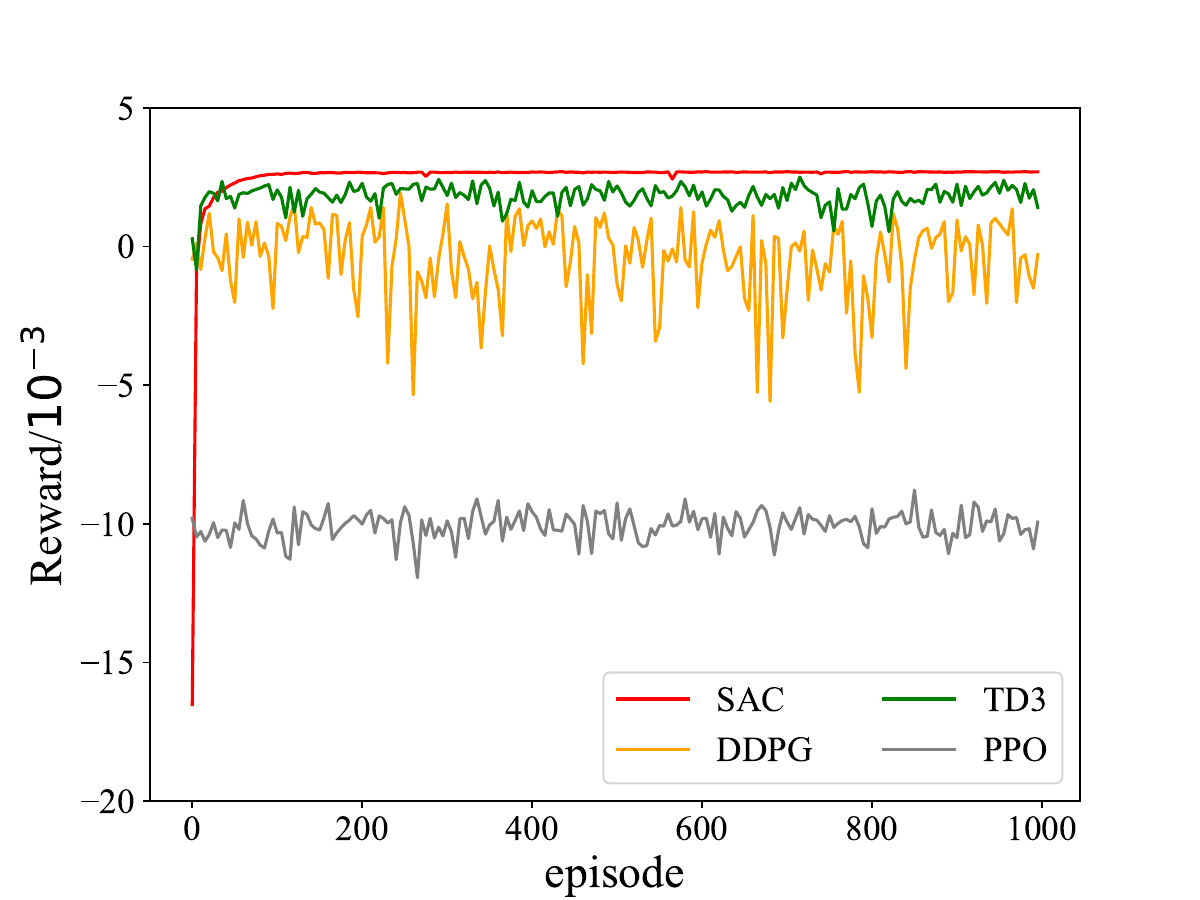}
		\label{fig14c}
	}
	\caption{Comparison between different RL algorithms}
	\label{fig14}
\end{figure*}
	
	All the above simulations assume that the images collected on the training vehicles are not affected by blurriness. Following, we will simulate the situations where some images appear motion blurred due to vehicle mobility. We assume that if the velocity of training vehicles exceeds $100 \text{km/h}$, $1/5$ of the collected image data will experience motion blur. Vehicles continue to use these images to train local models, which are then uploaded to BS for aggregation. 
	
	In Fig. \ref{fig10}, we adopt three aggregation methods, including DRL-BFSSL and two baseline algorithms, to aggregate for obtaining a global model with $1/5$ blurred images. Baseline1 represents BS uniformly utilizing the average federated to aggregate the parameters of local models. Baseline2 represents BS discarding local models trained by blurry image data, and then using average federated to aggregate the parameters of remaining local models. To validate the impact of blurry images on the model, we have set up baseline3, where $3/5$ of the images experience motion blur when the velocity exceeds  $100 \text{km/h}$. From the figure, it can be observed that the loss curves of all four methods exhibit a decreasing trend. 
	Based on the results of baseline1 and baseline2, it is evident that although baseline1 does not achieve a stable convergence of loss function, compared to baseline2, its loss function reaches a smaller value. In other words, because baseline2 discards some local models during aggregation, baseline2 uses the fewest number of local models among these methods. This suggests that the number of local models participating in aggregation is more critical than the quality of the models.	Finally, our proposed aggregation method of DRL-BFSSL not only reduces the fluctuation of the loss function but also converge to a smallest value above these methods. This is achieved by effectively addressing motion blur by assigning smaller weights to local models trained on blurred images, thereby reducing the negative impact of these models on the global model. From baseline3, it can be observed that an increasing number of blurry image data is unfavorable for the convergence of the loss function. However, compared to other aggregation methods, the convergence speed remains relatively stable.
	
	Increasing the number of local calculation times typically enhances classification accuracy; in other words, higher CPU frequencies often significantly improve classification performance. Therefore, in Fig. \ref{fig100}, we presented a comparative result of the utility between the SAC algorithm and traditional optimization methods based on heuristic and evolutionary algorithms, namely PSO, HGGA, and HGRA. To maximize resource utilization, lower average utility values reflect the superiority of an algorithm in resource allocation. As shown in the figure, SAC achieves the lowest utility within a reasonable range of CPU frequency selection. This indicates that SAC effectively balances computing power and energy consumption when optimizing resource allocation, thereby achieving higher classification performance. This advantage stems from its use of DRL strategies, which allow the algorithm more flexibly to the varying computational demands and resource constraints in dynamic environments, thereby enhancing overall performance.
	Additionally, in the inset of Fig. \ref{fig100}, we zoomed in the interval $[2850, 3100]$, which clearly illustrates that during the first 1000 episodes, the SAC algorithm rapidly reaches a local minimum utility, followed by a gradual increase and eventual stabilization. This shows that SAC quickly finds the highest CPU frequency within a reasonable range during the initial phase. However, as training progresses, the algorithm gradually discovers a balance point between computational performance and energy consumption, leading to a slight decrease in CPU frequency and eventual stabilization at this compromise point.

\begin{figure*}[!t]
	\centering
	\subfloat[Optimization problem Comparison]{%
		\includegraphics[width=0.32\linewidth]{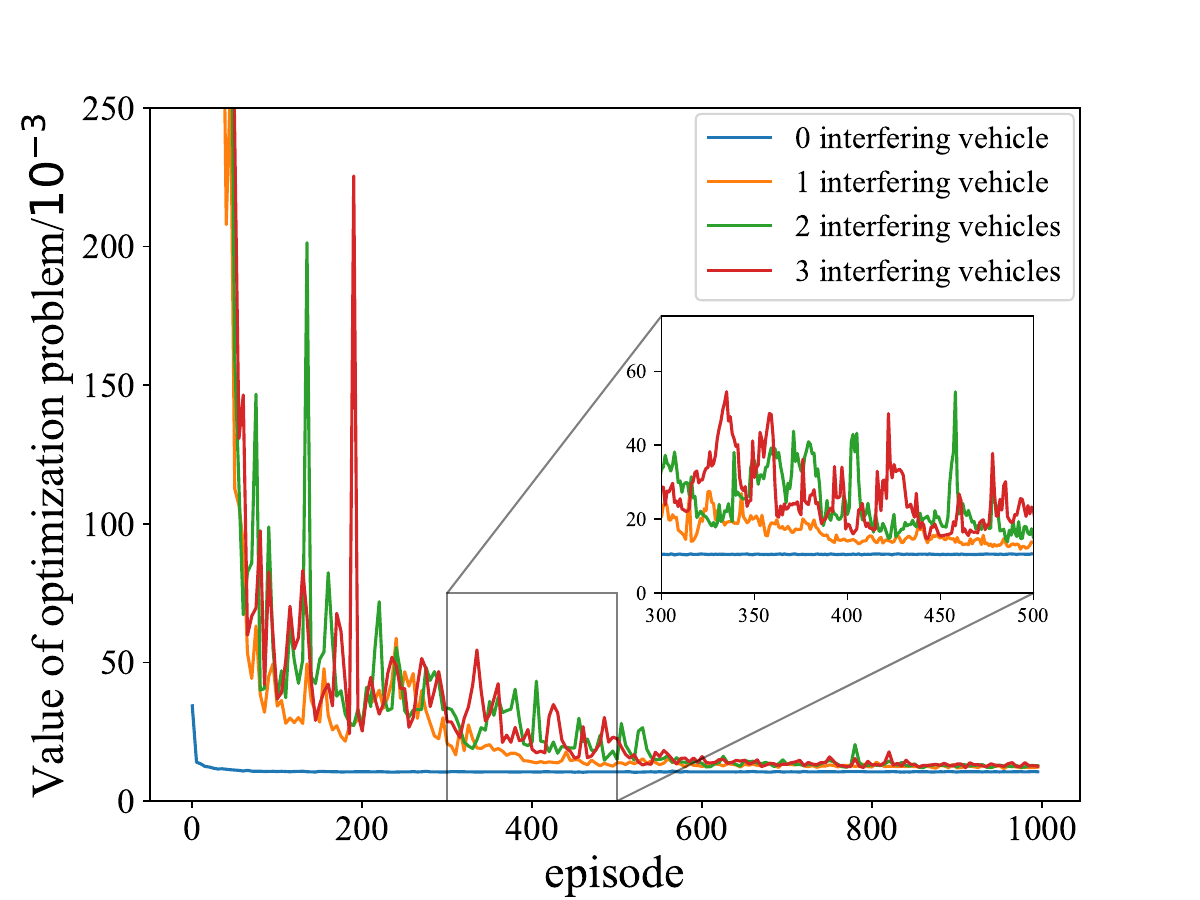}
		\label{fig16a}
	}\hfill
	\subfloat[Calculation Comparison]{%
		\includegraphics[width=0.32\linewidth]{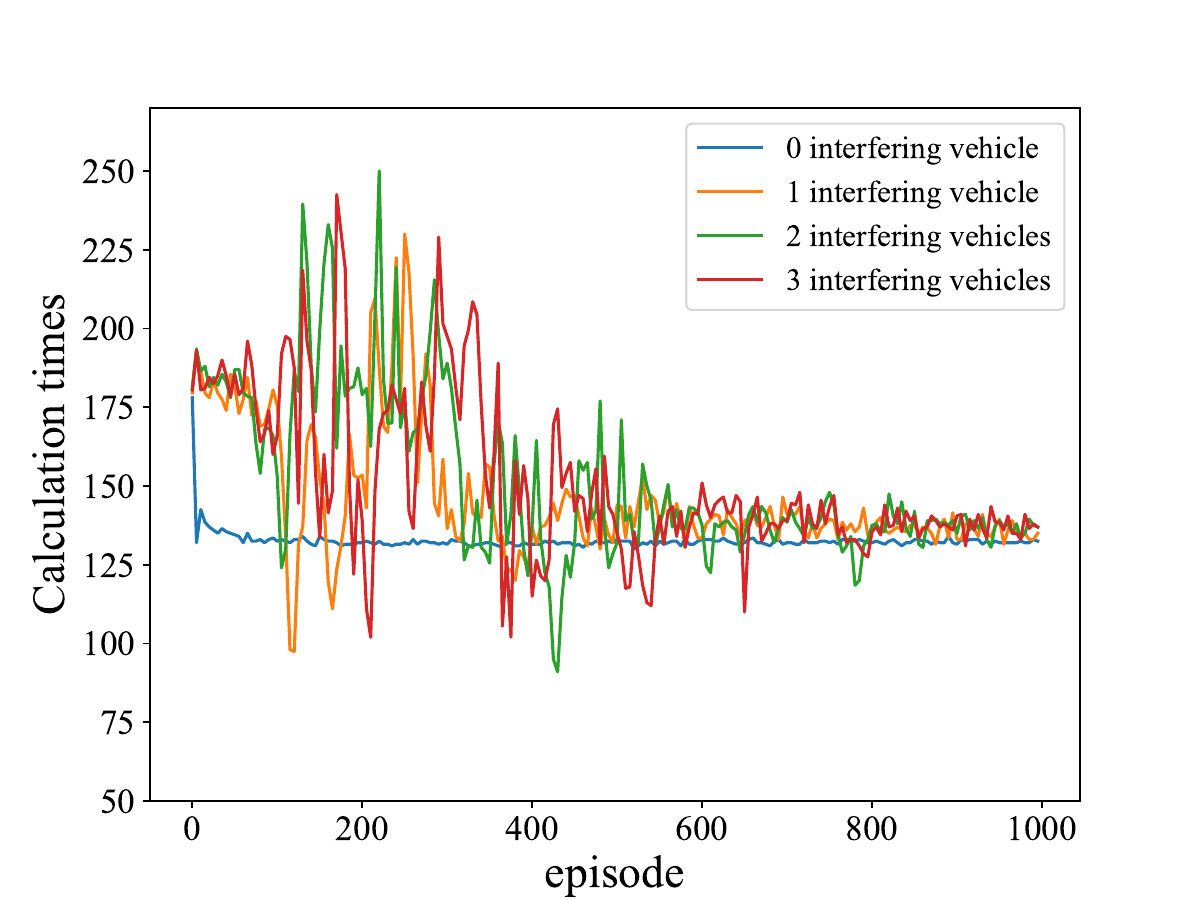}
		\label{fig16b}
	}\hfill
	\subfloat[Reward Comparison]{%
		\includegraphics[width=0.32\linewidth]{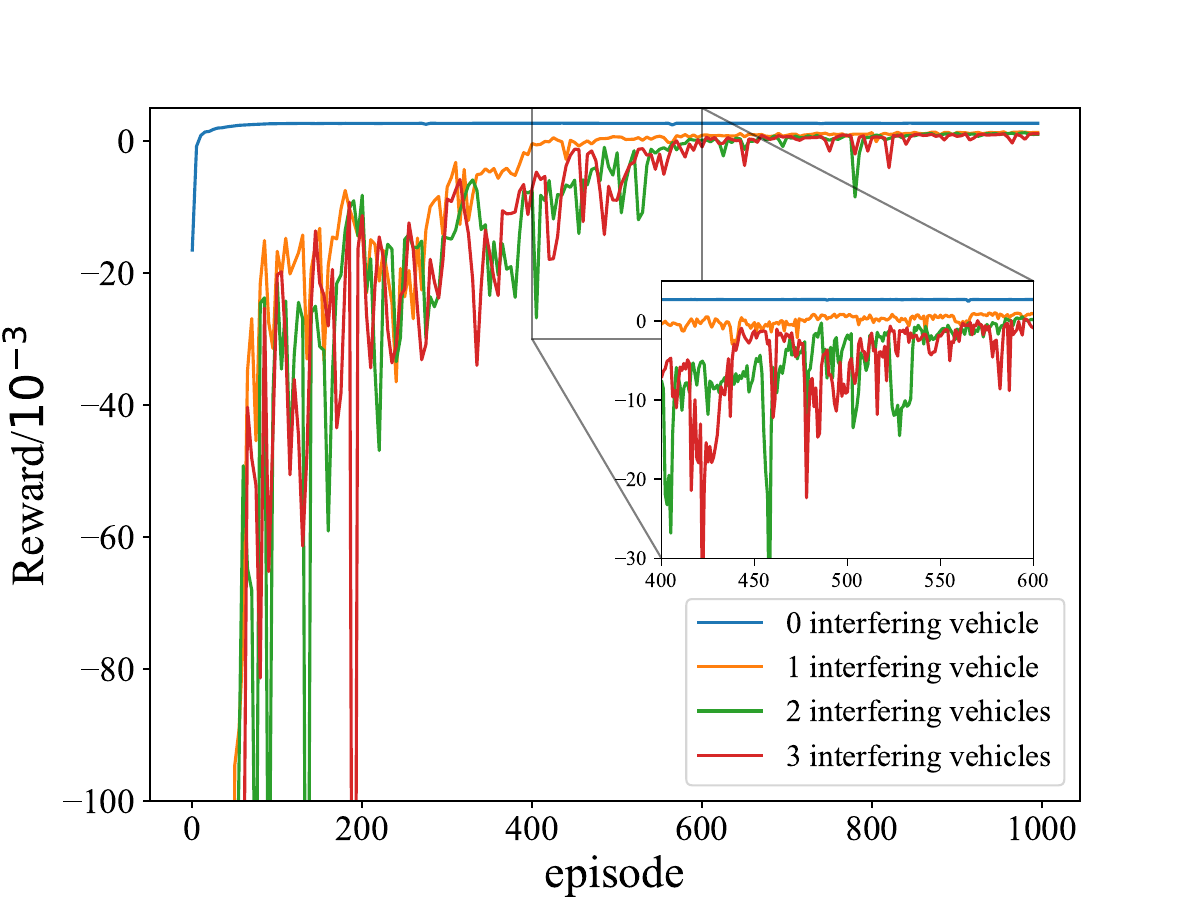}
		\label{fig16c}
	}
	\caption{Comparison between different number of interference vehicles}
	\label{fig16}
\end{figure*}

	In Fig. \ref{fig14}, we compared the performance of SAC, DDPG, TD3 and PPO algorithms in solving the optimization problem, covering the value of optimization problem, calculation times, and reward functions. From Fig. \ref{fig14a}, it is evident that the SAC algorithm rapidly reaches convergence and remains stable. DDPG and TD3 show similar performance, but TD3 slightly outperforming DDPG. The PPO algorithm exhibits a fluctuation level similar to TD3, but it fails to explore a suitable value of optimization problem. Overall, the SAC algorithm can obtain better solutions more quickly and more stably compared to other methods. This is mainly because SAC introduces entropy rewards to encourage policy exploration, maintaining a certain level of exploration during training.
	In Fig. \ref{fig14b}, we presented a comparison of calculation times. The calculation times obtained by PPO algorithm are high. The calculation times for the SAC, DDPG, and TD3 algorithms fluctuate around a certain value, with SAC having the smallest fluctuation, followed by TD3, and then DDPG.
	Fig. \ref{fig14c} shows the reward functions for each algorithm. The reward function of the SAC algorithm converges quickly and tends to be stable. The TD3 algorithm ultimately achieves a reward value that is second only to the SAC algorithm but exhibits some minor fluctuations. In contrast, the DDPG algorithm shows larger fluctuations in reward values and does not display a clear convergence trend. The PPO algorithm achieves relatively stable reward values, but its rewards are significantly lower than those of the SAC, DDPG, and TD3 algorithms. The SAC algorithm uses an entropy regularization strategy that better balances exploration and exploitation, allowing it to find the optimal solution more quickly and stably. The TD3 algorithm introduces instability in some scenarios through its dual policy delay update mechanism. Meanwhile, the DDPG algorithm is sensitive to hyper-parameters and lacks sufficient exploration mechanisms, making it prone to getting stuck in local optima and difficult to converge. PPO's policy updates are more conservative, and while it performs well in discrete action spaces, it does not achieve optimal performance in complex continuous action spaces.
	We aim to have sufficient calculation times to facilitate the training of the global model while minimizing the value of optimization problem. 
	From Fig. \ref{fig14b}, we can see that the SAC algorithm does not achieve the highest calculation times. However, as shown in Fig. \ref{fig14c}, the final reward value of the SAC algorithm is the highest among the four methods. This indicates that the SAC algorithm does not focus solely on maximizing the calculation times during training. Instead, it aims to achieve a relatively reasonable calculation times and pursue higher rewards.
	
	In Fig. \ref{fig16}, we compared the performance of SAC algorithm under different numbers of interfering vehicles, including the value of optimization problem, calculation times and the reward function. Overall, the results of the three can converge to relatively stable values.

	 Specifically, Fig. \ref{fig16a} and Fig. \ref{fig16b} illustrate the value of optimization problem and calculation times under SAC algorithm in the presence of varying numbers of interfering vehicles. When there are no interfering vehicles, the value of optimization problem and the calculation times converge to a minimum value and remains stable quickly. However, with the introduction of interfering vehicles, the value of optimization problem and calculation times noticeably increases, and gradually converging to a stable value. This is because when the number of interfering vehicles increases, the system needs to allocate more energy to the vehicles for training. This behavior of converging can be attributed to the SAC algorithm's ability to adapt to increased complexity and interference, effectively maintaining a robust performance even under more challenging conditions. Fig. \ref{fig16c} shows a comparison of the reward function throughout the process. In general, when there are no interfering vehicles, the reward function can quickly converge to a higher value. However, as the number of interfering vehicles increases, the reward function decreases. At the same time, it can be seen that as long as interfering vehicles are introduced, the overall system reward will significantly decrease. The increase in interfering vehicles makes management and energy allocation more complex, thereby reducing the efficiency and effectiveness of the SAC algorithm.
	
\begin{figure}
		\center
		\includegraphics[scale=0.38]{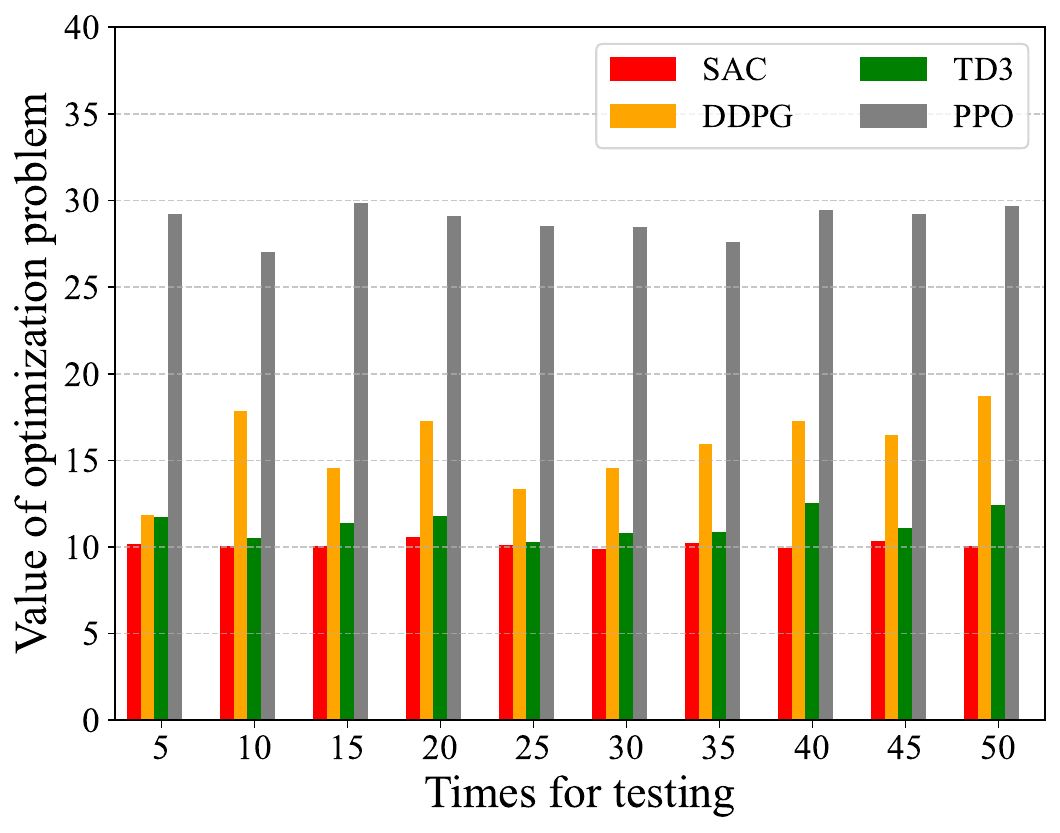}
		\caption{Testing results}
		\label{fig15}
	\end{figure}

	With the absence of interfering vehicles, we trained models using both the SAC, DDPG, TD3 and PPO algorithms and saved their parameters for testing. We conducted 50 tests and presented the results in Fig. \ref{fig15}. As shown in Fig. \ref{fig15}, the overall results indicate that the value of optimization problem produced by the SAC algorithm are smaller and more stable compared to those produced by the other algorithms. This suggests that the SAC algorithm outperforms the other algorithms in terms of both stability and performance. This superiority is likely due to the entropy regularization strategy employed by the SAC algorithm, which allows for a better balance between exploration and exploitation, leading to more stable convergence to superior solutions. In contrast, the DDPG and TD3 algorithm may be more prone to getting stuck in local optima during policy optimization, resulting in lower quality and less stable solutions.
	
	\section{Conclusions}
	\label{sec7}
		In this paper, we proposed DRL-BFSSL algorithm effectively mitigates motion blur while achieving high classification accuracy under privacy protection and label-free conditions. Additionally, the proposed resource allocation strategy based on the SAC algorithm minimizes system energy consumption and delay. Therefore, the method can be extended to various application fields, including traffic management and environmental monitoring in smart cities, resource allocation and scheduling in industrial automation, and enhancing system safety and stability in autonomous driving. The contributions of this paper can be summarized as follows:
	
	\begin{itemize}
		\item We proposed the DRL-BFSSL algorithm, which protects vehicle privacy, eliminates the need for labels in local training and achieves higher classification accuracy on both IID and Non-IID datasets after aggregating. 
		\item For the case of blurred images in IoV, our proposed aggregation method can effectively resist motion blur, resulting in a steady decrease in the loss function.
		\item Based on the SAC algorithm, our proposed resource allocation scheme minimizes energy consumption and delay while maintaining a stable number of calculation times and achieving high rewards. Even with interference, this algorithm still obtains high and stable rewards.
		
	\end{itemize}
	
	Although the DRL-BFSSL algorithm has demonstrated good performance in smaller-scale IoV environments, it may face scalability issues in larger or more complex networks. Additionally, this study only estimates motion blur from the perspective of vehicle velocity, and challenges remain in addressing sensitivity to motion blur caused by multiple factors, such as moving environments, whether as well as vehicle change directions etc. In the future, we will further investigate these two aspects to enhance the scalability of the algorithm and the accuracy of motion blur estimation.
	
	\begin{appendices}

		\section*{Appendix A} 		
		
		According to Eq. \eqref{eq33}, let $\varphi=\max{\left(\frac{E}{\beta_{k,t}^{n}}+\frac{F}{f_{k,t}^{n}}\right)}$, and Eq. \eqref{eq24} can be obtained:		
		
		\begin{equation}
			\begin{split}
				\underset{\boldsymbol{\beta}, \boldsymbol{p}, \boldsymbol{f}}{\min} \mathcal{C} : & \sum_{n=1}^{N}\left[\frac{A}{\beta_{k,t}^{n}}+B\left({f_{k,t}^{n}}\right)^3-C\left({f_{k,t}^{n}}\right)^2\right] +\varphi.
			\end{split}
			\label{eq34}
		\end{equation}
		
		\hspace{4em}\text{s.t.}
		\begin{subequations}
			\begin{equation}
				0 < \beta_{k,t}^{n} < 1\tag{\ref{eq34}{a}} \label{eq34a}
			\end{equation}
			\begin{equation}
				\sum_{n=1}^{N}\beta_{k,t}^{n} \le 1 \tag{\ref{eq34}{b}} \label{eq34b}
			\end{equation}
			\begin{equation}
				\epsilon_{k,t}^{n} \le \epsilon_\tau \tag{\ref{eq34}{c}} \label{eq34c}
			\end{equation}
			\begin{equation}
				p^{\text{min}} \le p_{k,t}^{n} \le p^{\text{max}}\tag{\ref{eq34}{d}} \label{eq34d}
			\end{equation}
			\begin{equation}
				f^{\text{min}} \le f_{k,t}^{n} \le f^{\text{max}}\tag{\ref{eq34}{e}} \label{eq34e}
			\end{equation}
			\begin{equation}
				\frac{E}{\beta_{k,t}^{n}}+\frac{F}{f_{k,t}^{n}}\le \varphi\tag{\ref{eq34}{f}} \label{eq34f}
			\end{equation}
			
		\end{subequations}		
		
		\textbf{\emph{Theorem1}}: The subproblem of the optimization problem \eqref{eq33} is a convex function.
		
		\textbf{\emph{Proof}}: The formula $\mathcal{C}$ consists of four parts: $\frac{A}{\beta_{k,t}^{n}}$, $B\left({f_{k,t}^{n}}\right)^3$, $C\left({f_{k,t}^{n}}\right)^2$ and $\varphi$. Since each part is a convex function and the constraints are all affine, so Eq. \eqref{eq33} is also a convex function.
		
		Using the Lagrange Multiplier Method to find the extreme values of a multivariable function subject to multiple constraints, the Lagrange equation can be expressed as:		
		
		\begin{equation}
			\begin{split}
				\mathcal{C} = & \sum_{n=1}^{N}\left[\frac{A}{\beta_{k,t}^{n}}+B\left({f_{k,t}^{n}}\right)^3-C\left({f_{k,t}^{n}}\right)^2\right] \\
				&+ \varphi +\phi\left(\sum_{n=1}^{N}\beta_{k,t}^{n}-1\right)	+\tau \left(\frac{E}{\beta_{k,t}^{n}}+\frac{F}{f_{k,t}^{n}}-\varphi\right),
			\end{split}
			\label{eq53}
		\end{equation}where $\phi$ and $\tau$ are Lagrange Multipliers associated with constraints \eqref{eq34b} and \eqref{eq34f}, utilizing the KKT conditions, we have the following equations:
		
		\begin{equation}
			\frac{\partial\mathcal{C}}{\partial\beta_{k,t}^{n}} = \phi - \frac{A + \tau E}{{\left(\beta_{k,t}^{n}\right)}^2} = 0,
			\label{eq54}
		\end{equation}
		\begin{equation}
			\frac{\partial\mathcal{C}}{\partial f_{k,t}^{n}} = 3B\left({f_{k,t}^{n}}\right)^2 - 2Cf_{k,t}^{n} - \frac{\tau F}{{\left({f_{k,t}^{n}}\right)^2}} = 0,
			\label{eq55}
		\end{equation}
		\begin{equation}
			\phi\left(\sum_{n=1}^{N}\beta_{k,t}^{n}-1\right)=0,
			\label{eq56}
		\end{equation}
		
		\begin{equation}
			\tau\left(\frac{E}{\beta_{k,t}^{n}}+\frac{F}{f_{k,t}^{n}}-\varphi\right)=0.
			\label{eq57}
		\end{equation}
		
		From Eq. \eqref{eq54} and Eq. \eqref{eq55}, we can derive out 
		
		\begin{equation}
			\phi=\frac{A+\tau E}{\left({\beta_{k,t}^{n}}\right)^2},
			\label{eq58}
		\end{equation}
		
		\begin{equation}
			\frac{3B\left({f_{k,t}^{n}}\right)^4-2C\left({f_{k,t}^{n}}\right)^3}{F}=\tau.
			\label{eq59}
		\end{equation}
		
		By transforming Eq. \eqref{eq58}, we can obtain:
		
		\begin{equation}
			\beta_{k,t}^{n}=\frac{\left(A+\tau E\right)^\frac{1}{2}}{\phi^\frac{1}{2}},
			\label{eq60}
		\end{equation}
		
		and from Eq. \eqref{eq56}, we can get
		
		\begin{equation}
			\phi\sum_{n=1}^{N}{\beta_{k,t}^{n}=\phi}.
			\label{eq61}
		\end{equation}
		
		Substitute Eq. \eqref{eq60} into Eq. \eqref{eq61}, we can obtain
		\begin{equation}
			\sum_{n=1}^{N}\left(A+\tau E\right)^\frac{1}{2}=\phi^\frac{1}{2},
			\label{eq62}
		\end{equation}
		
		Substitute Eq. \eqref{eq62} into Eq. \eqref{eq60}, the final result can be obtained:
		\begin{equation}
			\beta_{k,t}^{n}=\frac{{(A+\tau E)}^\frac{1}{2}}{\sum_{n=1}^{N}{(A+\tau E)}^\frac{1}{2}},
			\label{eq63}
		\end{equation}where
		
		\begin{equation}
			\tau=\frac{3B \left({f_{k,t}^{n}}\right)^4-2C\left({f_{k,t}^{n}}\right)^3}{F}.
			\label{eq64}
		\end{equation}		
		
	\end{appendices}

\end{document}